\documentclass{article}

    \PassOptionsToPackage{numbers}{natbib}

\usepackage[main, final]{neurips_2026}

\usepackage[utf8]{inputenc} %
\usepackage[T1]{fontenc}    %
\usepackage{hyperref}       %
\usepackage{url}            %
\usepackage{booktabs}       %
\usepackage{amsfonts}       %
\usepackage{amsmath}
\usepackage{nicefrac}       %
\usepackage{microtype}      %
\usepackage{xcolor}         %
\usepackage{graphicx}
\usepackage{multicol}
\usepackage{tocloft}
\usepackage{xstring}

\providecommand{\thanks}[1]{\footnote{#1}}

\let\oldparagraph\paragraph
\renewcommand{\paragraph}[1]{%
  \oldparagraph{#1}%
  \IfEndWith{#1}{.}%
    {\StrGobbleRight{#1}{1}[\stripped]}%
    {\def\stripped{#1}}%
  \addcontentsline{toc}{subsubsection}{\stripped}%
}
\title{Metacognition in LLMs:\\Foundations, Progress, and Opportunities}

\author{%
  \textbf{Gabrielle Kaili-May Liu}\textsuperscript{1}
  \qquad
  \textbf{Areeb Gani}\textsuperscript{1}$^*$
  \qquad
  \textbf{Jacqueline Lu}\textsuperscript{1}$^*$
  \qquad
  \textbf{Jordan Thomas}\textsuperscript{1}$^*$ \\
  \And
  \textbf{Mark Steyvers}\textsuperscript{2} 
  \qquad
  \textbf{Arman Cohan}\textsuperscript{1}  \\
  \\
  {\normalfont \textsuperscript{1}Yale University\quad\textsuperscript{2}University of California, Irvine}\\
  \\
  {\small \texttt{\{kaili.liu, arman.cohan\}@yale.edu}} \\
}

\begin{document}
\maketitle
{\renewcommand{\thefootnote}{*}\footnotetext{These authors contributed equally to this work.}}

\begin{abstract}
Metacognition is a foundational component of intelligence critical to effective learning, problem solving, decision-making, communication, and more. In recent years, it has become increasingly recognized as a cornerstone of capable, transparent AI systems. Yet while LLMs have made significant progress across diverse real-world tasks, it is not yet clear when, how, or to what extent they can exhibit or be endowed with effective metacognitive abilities, nor how such abilities can be adapted to advance the fundamental capabilities, reliability, and intelligence of AI systems. This paper bridges this gap by presenting the first comprehensive overview of the current state of knowledge on metacognition for LLMs. We analyze and taxonomize the landscape of this emerging field and summarize recent technical advancements, including methods and benchmarks to measure and evaluate LLMs' metacognitive abilities, techniques to elicit, improve, and apply metacognition in LLMs, and findings and implications of ongoing research. We also discuss applications, open questions and challenges, and promising directions for future work. Our aim is to provide a detailed and up-to-date review of this topic and stimulate meaningful research and discussion. An organized list of papers can be found at \url{https://github.com/yale-nlp/LLM-Metacognition}.

\end{abstract}

\begin{figure}[h!]
    \centering
    \includegraphics[width=.9\linewidth]{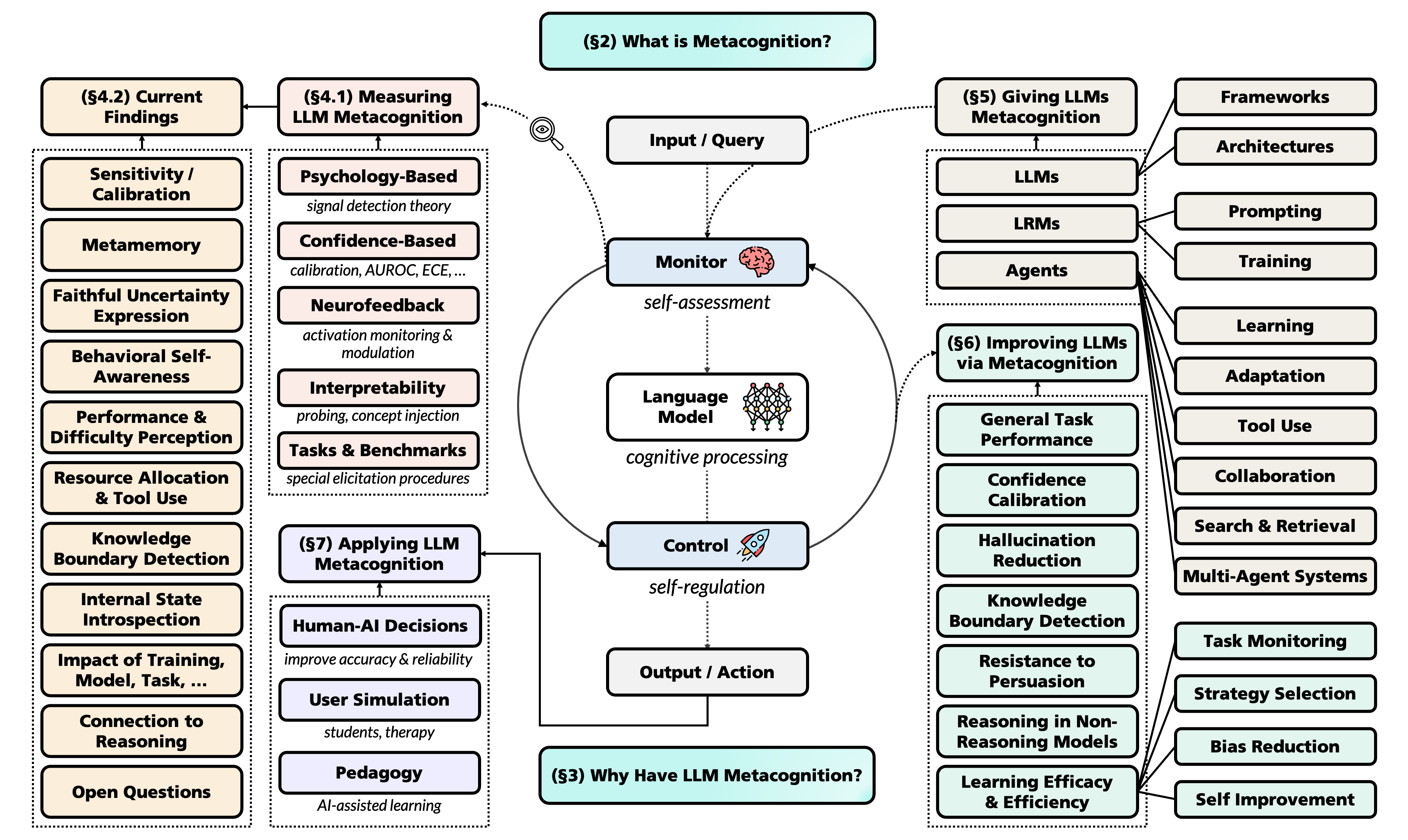}
    \caption{
    \textbf{Taxonomy of current research on metacognition in LLMs. }
    Metacognition describes the capacity for a system to assess and regulate its own cognition. The metacognitive loop consists of two interacting processes: \textit{monitoring} (e.g., forming judgments of uncertainty, task performance, or progress) and \textit{control} (e.g., planning, strategy selection, or effort re-allocation based on monitoring). 
    }
    \label{fig:fig1}
\end{figure}
\newpage

\renewcommand{\contentsname}{Contents}
\renewcommand{\cfttoctitlefont}{\Huge\bfseries}
\renewcommand{\cftaftertoctitle}{\vspace{1em}}

\renewcommand{\cftsecfont}{\bfseries}
\renewcommand{\cftsecpagefont}{\bfseries}
\setlength{\cftbeforesecskip}{0.8em}

\setlength{\cftsubsecindent}{1.8em}
\setlength{\cftsubsecnumwidth}{2.5em}
\setlength{\cftbeforesubsecskip}{0.25em}

\setlength{\cftparskip}{3.5pt}
\setlength{\cftsubsubsecindent}{5em}
\tableofcontents

\newpage
\section{Introduction}

Metacognition \citep{flavell1979metacognition, nelson1990metamemory, dunlosky2008metacognition} refers to the ability to monitor, assess, and regulate one’s own cognitive processes. It is crucial for learning, decision-making, and communication and plays a central role in continuous adaptation of behaviors and skills across diverse environments \citep{veenman2006metacognition, koriat2015metacognition, moshman2018metacognitive}. In humans, metacognition allows individuals to introspect and calibrate self-assessment of capabilities, choose suitable strategies to complete tasks, and optimize learning processes and task performance \citep{metcalfe1986feeling, koriat2007metacognitive, chen2017strategic}. This metacognitive flexibility makes reasoning robust to unseen problems, enables efficient problem-solving, and admits iterative, online learning \citep{ackerman2017meta, VALIENTE2026108131}. The study of metacognition is therefore important in fields of psychology, pedagogy, philosophy, and computer science.

Since metacognition is a hallmark of intelligence that is frequently considered missing in current AI systems \citep{schmill2008role, johnson2025imagining, von2025knowing}, an increasing number of studies have begun to draw connections between LLMs and metacognition. In the past few years, LLMs have made remarkable progress in natural language processing and beyond, driving advancements across high-stakes domains such as medical diagnosis \citep{Johnson2023AssessingTA, zhou2025large}, legal consulting \citep{dahl2024largelegalfictions, li-etal-2025-legalagentbench}, and scientific discovery \citep{song2025evaluatinglargelanguagemodels, zhang2025advancingscientificmethodlarge}. As LLMs increasingly mimic human cognitive faculties across diverse tasks \citep{tang2024humanlike, wang2024survey, geva2025llms}, a natural question arises as to whether such models can exhibit metacognitive behaviors---and whether endowing LLM-based systems with metacognitive abilities can improve their performance and capabilities, boost learning efficacy, and enhance outcomes and reliability in human-AI collaborative settings \citep{cappelen2024introspective, huff2025judgments}. It has been shown, for example, that prompting models to engage in metacognitive self-reflection can notably strengthen performance on reasoning tasks \citep{gao2024meta, balappanawar2025if, khandelwal2025language, liu2025instruct, qian2025cognitive, fang2026probabilistic} and improve the faithfulness of models’ verbalized uncertainty \citep{metafaith, rlmf}.  As metacognition underpins the ability to assess and communicate uncertainty in a transparent fashion, the extent to which it may be observed in LLMs is central to understanding their limitations and prospects for safer deployment \citep{steyvers2025improving}. Moreover, metacognition may present a natural pathway toward self-improvement \citep{anderson2005logic, liu2025position}.

Despite these promising directions, investigation and discussion of metacognition for LLMs remains relatively sparse and fragmented, with little unified direction or consensus on how to define, characterize, implement, and assess metacognitive capabilities in such systems. It remains unclear if LLMs can genuinely exhibit or mimic metacognitive behaviors, in what settings and to what extent they can do so, how post-training shapes metacognitive abilities, and how such abilities can be leveraged to advance the capabilities of LLMs. For example, studies such as \citet{ackerman2025evidence, chen2026learning, li2026language} suggest that 
LLMs can demonstrate reflective behaviors reminiscent of metacognitive processing, while works such as \citet{prasad2025two, cash2026quantifying, shatrade} conclude that LLMs are still far from being able to issue metacognitively effective confidence judgments. 
Such tension is echoed by \citet{steyvers2025metacognition}, who posit that important differences remain between human and LLM metacognition, including the extent to which metacognitive abilities can be improved through feedback and training and whether such skills are domain-general or task-specific \citep{fitzgerald2017domain, morales2018domain, carpenter2019domain}.

In this paper, we provide a comprehensive overview of metacognition for LLMs (Fig. \ref{fig:fig1}), presenting the first thorough and organized review of current research and knowledge on this emerging topic, with the aim of stimulating insightful discussion and future work. We initiate our exploration by defining and clarifying the concept of metacognition (\S\ref{sec:whatismetacognition}) and its importance (\S\ref{sec:importanceofmetacognition}). Subsequently, we turn our attention to methods and benchmarks for eliciting and evaluating metacognitive abilities of LLMs (\S\ref{subsec:measuringmetacognition}), the key findings and implications of this research (\S\ref{subsec:currentfindings}), techniques for instantiating and improving metacognitive skills in LLMs, LRMs, and agentic systems (\S\ref{sec:endowingmetacognition}), and metacognition-inspired methods to improve diverse LLM capabilities (\S\ref{sec:applyingmctoimprovecapabilities}). Finally, we outline applications of  LLM metacognition (\S\ref{sec:applications}), reflect on the current state of the field (\S\ref{sec:discussion}), and discuss future research directions (\S\ref{sec:futuredirections}). To the best of our knowledge, no prior work comprehensively and systematically examines the trends, technical advancements, and key findings on metacognition in LLMs. We thus aim to bridge this gap, shedding light on metacognition for LLMs as a critical yet underexplored facet that can spur progress toward more capable, reliable, and self-driven systems.

\section{What is Metacognition?} \label{sec:whatismetacognition}
Metacognition describes the ability to monitor and regulate one’s own cognitive processes. Although the term is used across scientific, philosophical, and pedagogical literature, each field emphasizes different aspects, applications, and implications. To help elucidate the concept, we summarize the core mechanism, components, and functions of metacognition which are commonly recognized in humans and of broader utility in the context of LLM-based intelligent systems.

\paragraph{Metacognition in Humans.} Metacognition
originates in cognitive psychology and has been extensively studied across developmental and educational settings \citep{flavell1979metacognition, metcalfe1986feeling, nelson1990metamemory, koriat2000conscious, koriat2007metacognitive, dunlosky2008metacognition, dunlosky2013handbook, koriat2015metacognition}. It is generally conceptualized as an internal perception-action loop consisting of self-assessment (\textit{metacognitive monitoring}) and self-regulation (\textit{metacognitive control}): individuals evaluate their competence on a task and use this assessment to guide strategic selection and control of subsequent actions. Within this framework, metacognition can be decomposed into several distinct, interacting components. \textit{Metacognitive knowledge} refers to awareness of one’s cognitive processes and capabilities (e.g., awareness of one’s own thinking patterns). It is subdivided into declarative, procedural, and conditional knowledge, which respectively capture what an individual knows about themself as a learner, how they apply cognitive strategies, and when and why specific strategies are appropriate. 
\textit{Metacognitive regulation} encompasses subsequent processes such as planning (e.g., setting goals, selecting appropriate strategies for a task)
and evaluation (e.g., reflecting on the effectiveness of applied strategies).
\textit{Metacognitive experience} refers to the subjective internal states---such as feelings of knowing, judgments of learning, or confidence---that arise during task execution (e.g., feeling enjoyment in solving a task). These signals support \textit{metacognitive judgments,} wherein one assesses their own performance before, during, or after a task \citep{schraw2009conceptual}. The efficacy of such judgments can be measured via \textit{metacognitive accuracy} (i.e., \textit{sensitivity})---how well one’s confidence can discriminate correct and incorrect responses---and \textit{metacognitive calibration}---how well one’s confidence reflects their probability of correctness or task accuracy, among other measures. Together, metacognitive knowledge and regulation support processes such as goal setting and decision-making.

Metacognitive strategies are distinct from cognitive strategies 
\citep{akturk2011literature} and critical to effective resource allocation and performance optimization \citep{mccormick2003metacognition, cohen2012importance, chen2017strategic}. The utility of metacognition is well-documented in pedagogical settings to be strongly correlated with academic performance \citep{isaacson2006metacognitive, young2008metacognitive}. While poor-performing students often overestimate their abilities and under-prepare, underconfident students may allocate effort inefficiently \citep{kruger1999unskilled, dunning2011dunning}; these patterns reflect systematic miscalibration that can be improved through targeted training or acquisition of more experience, even in children as young as three years of age \citep{schraw1998promoting, kramarski2003enhancing, chatzipanteli2014development}. Such findings may offer insights for building and enhancing metacognitive abilities in intelligent systems. Beyond this, some evidence suggests animals also possess metacognitive monitoring and use it to regulate learning, with further implications for learning in artificial systems \citep{josyula2009modeling}.

\paragraph{Metacognition in LLMs.} The concept of metacognition for language models has gained traction since 2023, but there is not a clear definition of what it entails. In the literature, “metacognition” is often used to describe straightforward reflection and (feedback-driven) revision of model outputs, particularly during reasoning tasks. However, such usage often only loosely aligns with principles of metacognition in psychological literature. More generally, consensus is lacking, and there is a wide variety of uses of the term, with different instantiations utilized depending on the setting of study \citep{li2025aiawareness}. For example, \citet{li2025adaptive} define metacognition as the ability to evaluate competence based on internal representations; \citet{lu2026auditing} define ``metacognitive confidence’’ as a model’s internal belief that it knows a claim regardless of factual correctness; \citet{lv2025whether} define ``metacognitive reasoning’’ as the assessment of whether a LM knows something;  \citet{bilal2025meta} define ``meta-thinking’’ as the combination of models’ self-reflection, assessment, and control of thinking; and \citet{fleig2025meta} define ``meta-reflection’’ as a model’s capacity to maintain performance by changing strategies in response to changing task constraints---where failure to do so would otherwise degrade performance---and require observable signatures of such adaptation as evidence. Many works also focus on ``meta-learning’’ processes inspired by metacognition, or propose metacognition-inspired methodologies with only loose or simplified interpretations of the concept \citep{wang2024metacognitive, li2025self, shan2025mentor, zeng2025mr}. For example, \citet{zeng2025mr} present a ``meta-reasoning’’ framework in which reasoning models are instructed to evaluate solution correctness, identify potential erroneous steps, and outline reasons for errors. In this paper, we adopt a broad view of metacognition to support our comprehensive examination and characterization of this emerging field. Thus, our discussion encompasses all such instantiations of metacognitive phenomena, while noting differences in conceptual grounding and operationalization.

\section{Why is Metacognition in LLMs Important?} \label{sec:importanceofmetacognition} 
In AI systems, metacognitive-like processes may manifest as models learn and optimize performance, raising fundamental questions about the nature, extent, and downstream impacts of metacognition in such systems in comparison to human metacognitive processing. Various works have highlighted the potential benefits of endowing intelligent systems with stronger metacognitive capacities \citep{schmill2008role, wei2024metacognitive, bergamaschi2025fast, lee2025metacognitive, shakarian2026toward}; in this section, we present a similar characterization of such benefits for LLMs, synthesizing observations across the literature to highlight the diverse array of fundamental capabilities and system-level benefits directly supported by improved metacognition.

\paragraph{Task Performance and Learning.} Since metacognitive processes support cognitive outcomes, encoding metacognitive abilities into LLMs (or integrating modules to execute metacognitive functions into LLM-based systems) can support improved task performance, boost learning efficacy and efficiency, and drive human-like behaviors and decision-making \citep{iqbal2023ai}. Analysis of large reasoning models (LRMs), for example, has revealed that such models often partake in reflective behaviors associated with metacognition \citep{gandhi2025cognitive, liu2025there, wen2025thinkpatterns}, with these thinking patterns supporting increased accuracy and reasoning trace complexity. 
Enabling models to monitor their own problem-solving processes can help them to validate selected strategies, recognize potential biases, identify opportunities for improvement, and assess confidence levels \citep{Bose2025}. Improved self-awareness of capabilities and limitations can strengthen models’ ability to request assistance when appropriate and facilitate self-regulated learning. Metacognitive faculties such as the perception and communication of uncertainty and knowledge boundaries \citep{fleur2021metacognition, ccini2023multiple} are likewise important in many LLM applications and can help models better detect and abstain in response to questions that are unanswerable or beyond the scope of their knowledge \citep{sikora2026}. 
These behaviors are directly supported by good metacognitive sensitivity and metacognitive calibration, which enable better confidence expression. The combination of cognitive and metacognitive functions has further been shown to improve decision quality while reducing resource consumption and unlock human-like adaptive learning, skill use, and cognitive control \citep{bergamaschi2025fast}.

\paragraph{Reliability and Trustworthiness.} Effective metacognitive monitoring is important to improving the transparency, reliability, and downstream utility of LLM-based systems, particularly in human-AI interaction settings. Measures of metacognitive sensitivity can help users to calibrate their reliance on model outputs and better incorporate external advice during AI-assisted decision-making \citep{lee2025metacognitive}. 
To earn human trust, LLMs must be able to accurately assess the likelihood of their predictions being correct \citep{steyvers2025large} and faithfully communicate such confidence or uncertainty to users \citep{yona2024can, metafaith}. 
For example, in clinical settings, AI systems may need to say ``I don't know'' when evidence is conflicting, key patient information is missing, or a question falls outside the system’s scope, thereby signaling uncertainty and directing the case toward clinician review rather than issuing a confident but unsupported response \citep{sikora2026}.
Consistent with this, better metacognitive sensitivity and faithful confidence expression have been shown to significantly enhance users’ perception of model accuracy \citep{steyvers2025metacognition, steyvers2025large}. These are further crucial to closing the gap between what LLMs know and what users believe they know \citep{steyvers2025large}, especially as LLMs are increasingly used for real-world decision-making.

\paragraph{Human-AI Collaboration.} Metacognition is also important to effective human-AI collaboration. In conversational settings, metacognition can enable models to acquire improved contextual self-awareness and leverage knowledge of their role in a conversation to better recognize when clarification is needed, detect inconsistencies, and adapt responses appropriately \citep{yun2026does}. In agentic settings (e.g., scientific reasoning), metacognition can improve agents’ ability to dynamically weigh the relative costs of different actions (e.g., retrieval vs. strategy evolution), strategically execute self-correction, and iteratively refine tool use \citep{lu2026beyond}. Strong metacognitive monitoring can further advance the utility of LLM-based systems in pedagogical applications (e.g., by assisting development of personalized learning plans) \citep{huff2025judgments}.

\paragraph{Hallucination Reduction.} Metacognition can play a role in mitigating hallucinations \citep{yona2026hallucinationsunderminetrustmetacognition}. In humans, discrepancies between performance and confidence level can predict hallucinatory or erroneous decisions and behavior \citep{wright2024experiencing}. Analogous findings have been raised for LLMs: \citet{simhi2025trust}, for example, show that LLMs engage in high-certainty hallucinations consistently across models and tasks, while \citet{lu2026auditing} demonstrate that LRMs likewise hallucinate with high confidence and reinforce biases and errors through flawed reflective processes which further the occurrence of hallucination. In fact, hallucinations by LLMs can often be characterized by specific metacognitive failure modes, including flaw repetition (e.g., looping, incorrect reasoning), think–answer mismatch (e.g., contradicting reasoning), and overconfidence (i.e., poor metacognitive calibration) \citep{yao2025reasoning, lu2026auditing}. Improving LLMs’ metacognitive capacities therefore presents an avenue to directly target the source of such limitations, including through more faithful verbalization of internal processes---which itself is a metacognitive function amenable to improvement.

\paragraph{Interpretability.} Lastly, metacognition can enhance model interpretability. Models that can self-identify errors, perform self-correction, and explain their correction procedure provide more transparent and interpretable outputs \citep{tan2025tuning}. More generally, rather than relying on complex procedures to analyze model internals, this presents the possibility of simply querying models to explicitly report their beliefs, goals, and thought processes.\footnote{Nonetheless, improved capacity for metacognitive monitoring and regulation may enable systems to strategically modify outputs or internal signals to evade oversight or pursue unintended objectives (e.g., \citet{li2026language}); we discuss such risks further in \S\ref{sec:discussion}.}.

\section{Do LLMs Have Metacognition?} \label{sec:dollmshavemetacognition}
Prior to conferring LLMs with improved metacognitive capabilities, it is desirable to first understand when and to what extent models are able to engage in metacognitive behaviors. To this end, several distinct strains of work to measure and benchmark metacognition in LLMs have emerged (\S\ref{subsec:measuringmetacognition}), in addition to numerous disjoint studies of specific metacognition-derived faculties of LLMs (\S\ref{subsec:currentfindings}).

\subsection{Measuring Metacognition in LLMs} \label{subsec:measuringmetacognition}
We introduce existing metrics and experimental procedures to measure and elicit metacognition in LLMs.

\paragraph{Psychologically-Grounded Measures.} In cognitive psychology, confidence is considered an overt indication of behavioral uncertainty whose alignment with task performance is well-established as a measurable indicator of metacognition \citep{fleming2012neural, fleming2012metacognition, fleming2017self, fleming2024metacognition}. A prominent paradigm for quantifying human metacognition is based on signal detection theory (SDT) \citep{maniscalco2012signal, barrett2013measures, fleming2014measure}, and it is designed to isolate metacognitive (“type 2”) sensitivity (the efficacy with which confidence ratings distinguish correct and incorrect responses) from confounding factors of response bias (tendency to be confident) and task performance (“type 1” sensitivity). Under this framework, meta-$d'$ and $d'$ are measures of metacognitive sensitivity and cognitive ability, respectively, and their ratio $\frac{\text{meta-}d'}{d'}$ (M-ratio) and difference meta-$d' - d'$ (M-diff) represent metacognitive efficiency by normalizing metacognitive sensitivity relative to type-1 task sensitivity.\footnote{An M-ratio of 1 indicates that an individual is an optimal
metacognitive observer whose confidence captures all the information available from type 1 evidence. On the other hand, $M<1$ indicates metacognitive loss, wherein the confidence signal is less informative than expected based on the evidence, and $M>1$ indicates the confidence signal assesses information beyond what is driving the observer’s type 1 judgments.} Recent work has also proposed information-theoretic alternatives to SDT, such as meta-$\mathcal{I}$ and relative metainformation, which quantify how much information confidence ratings carry about response accuracy while reducing reliance on parametric SDT assumptions \citep{dayan2023metacognitive, meyen2025information}.

Several works have proposed methodologies inspired by SDT to quantify the metacognitive ability of LLMs. Such approaches generally pair meta-$d'$ with a specific procedure to elicit model responses and associated confidence scores \citep{servajean2026measuring}. For example, \citet{phammetacognitive} use meta-$d'$ to quantify how reliably a model’s confidence (estimated as the maximum softmax output probability) predicts its own accuracy to inform test-time model selection, and \citet{park2026fine} utilize a dual-questioning protocol to elicit answers and self-evaluations of knowledge from LLMs to estimate meta-$d'$. \citet{DMC} adapt meta-$d'$/$d'$ logic to LLMs with Decoupling Metacognition from Cognition (DMC), in their experiments converting multiple-choice items into binary-choice tasks and eliciting answer confidence. DMC estimates task sensitivity $\hat d$ and computes $MC=d^*/\hat d$, making it closely analogous to the M-ratio while aiming to separate confidence-based failure prediction from raw task performance. 
Other psychology-based methodologies include the use of judgments of learning (i.e., self-predictions of future memory performance) \citep{huff2025judgments} or vectorized dimensional measurements of metacognitive capabilities of LLMs \citep{sethi2025llms}. While these approaches represent meaningful progress in measuring LLM metacognition, they are not without their limitations, including dependence on task-specific design or constrained task formats, sensitivity to confidence elicitation method, and lack of empirical justification.
Moreover, SDT-based approaches typically require constrained response formats or external correctness judgments, making their extension to open-ended generation less direct.

\paragraph{Neurofeedback-Based Measures.} Neurofeedback is an established neuroscientific technique in which participants are tasked with regulating their brain function in response to real-time biological feedback signals \citep{sitaram2017closed}. A classical example is in fear reduction experiments: subjects are presented with fear-inducing images, provided with a fear score derived from real-time recordings of their brain activity, and asked to modulate their neural activity to reduce the neurofeedback score. In a similar fashion, select works have sought to simulate a neurofeedback environment for LLMs to study their ability to monitor and control internal activations. This presents an alternative metacognitive evaluation route that does not depend on behavioral outputs (e.g., self-reported confidence scores may misrepresent actual uncertainty of a model). For example, \citet{li2026language} devise a neurofeedback approach based on in-context learning wherein models participate in a multi-turn dialogue and are tasked with either implicitly or explicitly modulating\footnote{The ability for models to report on neural activations is also explored.} internal activations, while \citet{yalon2026indications} explore a similar setup to test models’ ability to metacognitively predict the dominance of internal beliefs. Similar to psychologically-inspired methods, neurofeedback-based methods must also disentangle metacognitive from cognitive processing.

\paragraph{Confidence-Based Measures.} Another class of approach directly uses the quality of judgments of intrinsic confidence as a proxy for LLM metacognition. Such works mainly focus on large reasoning models: \citet{ma2025large} inspect the alignment between step-wise intrinsic confidence and process-level quality labels, quantified via typical calibration measures such as area under the receiver operating characteristic curve (AUROC) \citep{boyd2013area} and area under the precision-recall curve (AUPR) \citep{manning1999foundations}, while  \citet{lu2026auditing} define metacognitive confidence to estimate a model’s belief about its own knowledge, which may be miscalibrated with factuality and help to explain high-confidence hallucinations. These methods are only loosely based in psychology and do not discuss potential bias relative to type 1 task performance.

\paragraph{Interpretability-Based Measures.} As metacognition concerns the ability to introspect on internal states, other works have turned to interpretability techniques to detect signatures of metacognition in LLMs. This includes use of task-specific metacognitive probes to investigate whether correct and incorrect responses or reasoning traces are associated with statistically distinguishable internal computational signals (e.g., last token representations, self-attention scores, fully-connected activations, final layer softmax probabilities) \citep{li2025adaptive, li2026towards} and measurement of models’ ability to accurately detect the presence of injected concepts \citep{lindsey2026emergent}. While such analysis depends on access to model weights, it is generally extensible across models and task settings.

\paragraph{Task-Specific Measures.} Metacognition is a fundamental capability that operates within and across domains, but in the early pursuit of eliciting and understanding metacognition in LLMs, it can be advantageous to focus instead on targeted, task-specific evaluations.  To this end, some studies have designed novel experimental procedures to examine metacognitive capabilities of models in controlled settings. For example, \citet{ackerman2025evidence} introduce two experimental paradigms inspired by research on metacognition in nonhuman animals, wherein the ability for models to strategically deploy knowledge of internal states is tested without relying on self-reports; this includes assessment of models’ ability to identify their knowledge (i.e., do models know \textit{what} they know) and of models’ certainty of their knowledge (i.e., do models know \textit{that} they know), in a game-like setup. To evaluate LLM metacognition in dynamic contexts more reflective of real-world applications, \citet{prasad2025two} utilize a zero-sum multi-turn debate setup in which models are required to update beliefs and perform accurate self-assessment in extended interactions. Other efforts focus on investigation of specific metacognition-inspired capabilities, including selection and organization of task-specific skills \citep{didolkar2024metacognitive}, self-detection of biases during essay writing \citep{Uzwyshyn2024}, reflection on creative generation tasks \citep{comsa2025does}, self-prediction of generation temperature, and metalinguistic analysis \citep{begus2025large}. In general, such approaches rely on carefully crafted prompts to elicit and inspect models’ higher-order metacognitive abilities.

\paragraph{Benchmarks for LLM Metacognition.} Beyond individual experimental procedures, it is possible to construct standardized benchmarks to more holistically evaluate the metacognitive competence of LLMs. Several recent benchmarks measure metacognition in LLMs by evaluating their ability to reason about their own knowledge, uncertainty, or intentions \citep{cacioli2026metacognitivebattery, naphade2026me}. CogEdit \citep{fan2026towards} evaluates metacognitive behavior during knowledge editing by constructing questions involving counterfactual, boundary-constrained (appropriate generalization without overgeneralizing), and noisy editing (filtering potentially unreliable information) scenarios, to capture self-awareness and reflective reasoning during model updates. MetaMedQA \citep{griot2025large} tests whether LLMs can recognize knowledge limitations and detect unanswerable questions in medical QA settings, by tasking models with making such metacognitive judgments on multiple-choice questions containing fictional concepts, malformed questions, and altered answers. ObjexMT \citep{kim2025objexmt} measures metacognitive calibration by asking models to infer the latent objective of multi-turn conversations and report confidence in that answer, with calibration measured via ECE and high-confidence error rates. Finally, AwareXtend \citep{passaro2025awareness} evaluates self-awareness and social awareness of LLMs across five dimensions of capability, mission, emotion, culture, and perspective, by observing performance on multiple-choice questions testing these abilities, and having several LLMs collaborate via voting or debate to produce the final answer. Together, these illustrate diverse aspects of metacognitive behavior that can be benchmarked in LLMs. Nevertheless, they remain constrained to specific task settings, disjoint in the types of metacognitive skills assessed, and subject to influence of general ability.

\paragraph{Gaps in Metacognitive Evaluation.} To date, higher order skills such as metacognition and creativity remain under-represented among LLM benchmarks, which tend to focus instead on evaluating lower-level cognitive capacities \citep{huber2025llms, zhang2025remembering}. This is partly driven by the difficulty of reliably and consistently measuring metacognitive performance, evidenced by the earlier-named limitations of the methodologies introduced above. Moreover, some analysis \citep{cacioli2026llmsassignaldetectors} shows that single-point confidence probes and post-hoc calibration techniques typically used in metacognitive assessment of LLMs exhibit consistent failure modes. This can potentially be explained by the impact of post-training procedures such as reinforcement learning with human feedback (RLHF), which can systematically alter models' internal uncertainty (e.g., change the underlying probability distributions) and thereby influence the connection to models’ output behavior \citep{zhou2023navigating}. Despite this, the incorporation of metacognitive evaluations into future benchmarks is important, as it can enable more accurate assessment and enhancement of model capabilities and inform oversight and safety. Especially as LLMs assume greater agency in real-world applications, the observation of consistent, stable, reproducible metacognitive phenomena in LLMs in certain settings (\S\ref{subsec:currentfindings}) warrants further investigation and inquiry. This motivates the development of more robust and systematic methods to measure, elicit, and evaluate metacognitive capabilities of LLMs across diverse contexts.

It is worth noting that while LLMs may exhibit behaviors that suggest a form of ``artificial metacognition’’ (e.g., assigning calibrated confidence scores to outputs, adapting outputs based on reflective feedback, performing self-referential inquiry), the autoregressive nature of such models has led to the view that they present no real (meta-)cognition to speak of \citep{huber2025llms}, with apparent metacognitive tendencies simply arising as a byproduct of LLMs’ replication of complex behavioral patterns derived from their training data \citep{szeider2025llm}. To this end, some have argued for more formal comparisons between human and LLM metacognitive processes \citep{pavlovic2024generative}; this can help to clarify whether models are simply simulating superficial aspects of natural metacognition, and inform the development of more human-like and effective intelligent systems.

\subsection{Current Findings on Metacognition in LLMs} \label{subsec:currentfindings}
While LLMs have shown great proficiency in various cognitive tasks, their ability to engage in metacognitive processes remains underexplored. In this section, we summarize the key findings and implications of current studies on metacognition in LLMs. We organize our discussion according to the type of metacognitive ability explored, roughly sorted from general to specific, and follow it by discussing the impact of factors such as model size, post-training, and confidence estimation method.

\paragraph{Metacognitive Sensitivity and Calibration.} Given the nascent state of research on LLM metacognition, a number of studies have aimed to simply profile the baseline metacognitive performance of LLMs. Beyond the findings of measurement-focused works such as those discussed in \S\ref{subsec:measuringmetacognition}, which establish that metacognitive sensitivity and efficiency of models are generally weak to moderate, analysis of proprietary LLMs \citep{pavlovic2024generative, cash2026quantifying} suggests that such models can, in fact, achieve better metacognitive sensitivity than humans on specific tasks (e.g., predicting outcomes of American football games), although such experiments tend to use small sample sizes (e.g., $<50$ examples), limiting generalizability. Findings on metacognitive calibration are mixed and task-dependent: while \citet{pavlovic2024generative} show models tend to be underconfident on a coaching task, \citet{cash2026quantifying} find across several tasks that LLMs are largely overconfident---a phenomenon well-documented\footnote{Systematic overconfidence is generally observed in a task-agnostic fashion in smaller models, and in difficult task settings in larger models. It can be attributed to a number of factors, such as RLHF post-training \citep{leng2025taming}, task length \citep{liu2025understanding}, and use of imbalanced datasets in which failure cases or uncertainty rarely occur \citep{zhou2023navigating, chung2025learning, stechly2025beyond} in the calibration literature \citep{wen2024human, xiong2024can, chhikara2025mind}}---and struggle to retrospectively adjust confidence judgments based on past performance, demonstrating a metacognitive limitation not observed in humans. Metacognitive deficiencies of open-source models are likewise apparent, with \citet{prasad2025two} identifying five consistent metacognitive failures of leading open-source and proprietary LLMs in a selection of debate-focused and multi-turn task settings, including initial overconfidence, escalating certainty, mutually impossible high confidence, self-debate bias, and misaligned private reasoning. \citet{shatrade} further show that metacognitive sensitivity can be harmed by reasoning, with long reasoning traces hindering accurate self-assessment despite boosting performance, and distillation of large models’ reasoning traces into smaller LMs impairing metacognitive sensitivity. Finally, \citet{li2024meta} examine the impact of declarative and procedural knowledge on model performance in a range of task settings, as these directly inform metacognitive processing in humans. They find that declarative information is generally more beneficial, while procedural knowledge is more useful in simple logical reasoning tasks, and that models’ ability to leverage both types improves during pre-training at rates that vary with training step and model size. Overall, while LLMs can somewhat successfully demonstrate functional metacognition, substantial gaps remain, presenting potential risks in downstream agentic and decision-making applications.

\paragraph{Metamemory.} The ability to introspect on memory is another core aspect of metacognition which has received attention \citep{liang2026metamemory}. \citet{huff2024metacognitive, huff2025judgments} study the ability for GPT models to predict their own future memory performance, known as judgments of learning (JOL) in humans. They find that models largely fail at this task, signaling a lack of associated monitoring mechanisms, and underscore the need for further advancements in LLMs’ self-monitoring capabilities, particularly to assist pedagogical applications. However, it is unclear whether such limitations extend to non-GPT models or more recent GPT variants.

\paragraph{Faithful Communication of Uncertainty and Decisions.} 
Another promising direction of metacognitive study aims to understand the extent to which models can understand and faithfully explain their own internal processes. For example, the ability to detect different sources of uncertainty at play when responding to a query, as well as the decision-making processes involved, and accurately communicate these to a user is important and aided by metacognitive awareness. To this end, several works \cite{yona2024can, metafaith} have turned to study the faithfulness with which LLMs express their intrinsic uncertainty—also known as \textit{faithful calibration}. The aim is to align models’ expressed and intrinsic uncertainty, where expressed uncertainty may be formulated numerically (e.g., ``90\% confident'') or embedded into the model’s output via use of linguistic uncertainty markers (e.g., ``It is likely that...'').\footnote{Faithful calibration thus differs from the traditional view of calibration which instead seeks to align confidence with accuracy; the two calibration dimensions may diverge, e.g., when a model's internal uncertainty deviates from its accuracy.} \citet{yona2024can, metafaith, mics, rlmf} recently demonstrated that both open- and closed-source frontier models largely fail to faithfully express their intrinsic uncertainty, even with the application of specialized prompts or existing calibration approaches. Despite this, metacognitive prompting---wherein a system prompt is used to instruct models that they possess good metacognition---and the use of \textit{metacognitive accuracy} as an \textit{additional} training signal during RL were found to consistently and significantly improve faithful calibration of diverse LLMs, and the quality and helpfulness of models’ resulting uncertainty expressions. This suggests that enabling models to better assess their own task performance can be important in expanding model capabilities. On the other hand, \citet{PPR:PPR1118147} study the ability for LLMs to explain their reasoning processes and the weight assigned to different factors in decision-making. They find that frontier LLMs such as GPT-5 can, in fact, hold and express metacognitive knowledge of their decision processes to a degree comparable to humans, but that this is a relatively new capability not observed in earlier versions such as GPT-4o. As faithful explanations measurably influence user trust in AI applications, enabling models to transparently disclose internal processes is central to appropriately calibrating user reliance on LLM outputs \citep{steyvers2025large}.

\paragraph{Self-Prediction of Task Performance and Difficulty.} Beyond contributing to transparency, metacognitive monitoring enables humans to self-assess performance and predict task difficulty, leading to more efficient and effective problem-solving via subsequent metacognitive regulation. Existing research suggests that while LLMs can engage in similar monitoring behaviors, the efficacy of such judgments is limited and generally does not translate to reliable strategic adaptation \citep{cao2026llmsknowwhentheyknow}. For example, \citet{barkan2025large} demonstrate that while leading LLMs can predict task success with better-than-random discriminatory power, such predictions do not improve with task familiarity, nor do reasoning models outperform non-reasoning LMs in making such judgments. \citet{hwang2025can} show that reasoning ability does not correlate with such metacognitive awareness, even when estimating the complexity of simple reading comprehension tasks. Analysis by \citet{binder2025looking, li2026towards} further reveals that prediction of task difficulty is particularly challenging for models in hard or out-of-distribution settings,\footnote{Despite this, in certain settings it appears that models can effectively assess and utilize their own confidence in answering factual and reasoning questions---for example in a synthetic game setup designed by \citet{ackerman2025evidence}. This provides additional evidence suggesting such self-assessment ability may be task-specific.} and that even when predictive metacognitive information is present, it often fails to lead to effective monitoring and behavior adaptation. Lastly, overconfidence was found to lead to poor decision-making in agentic settings, providing evidence for the negative impact of models’ poor awareness of their capabilities on task performance. This motivates the acquisition of a deeper understanding of models' self-assessment capabilities. Yet it is also important to consider potential risks of improved self-monitoring, including potential subversion of oversight mechanisms (\S\ref{sec:discussion}).

\paragraph{Resource Allocation and Tool Use.} Metacognition supports efficient allocation of computational and informational resources during learning and task execution. Yet LLMs lack the ability to reason about how to best allocate limited effort across multiple problems or options, including limited ability to estimate potential returns on allocated effort or to decide when to persist (e.g., continue searching) or move on \citep{zhao2026roi}. While \citet{li2025adaptive} show that a probe trained using metacognitive information can improve tool use decisions, this relies on external interventions rather than judgments by a model itself. Work in this sub-area is limited, and further investigation can be worthwhile, especially toward the development of more efficient agentic systems and effective tool use.

\paragraph{Awareness of Knowledge Limitations.} As it is generally desirable for a system to be able to recognize the scope of its own knowledge---to avoid misalignment between perceived and actual capabilities, improve reliability, and enable adaptive tool use or abstention---another line of research has sought to profile the degree to which LLMs embody this metacognitive attribute \citep{prato2024large}. Findings in this area are generally mixed, with some reporting a lack of such ability in LLMs \citep{yin2023large, griot2025large, kale2025line, wang2026mirror}, and others suggesting models can exhibit functional access to knowledge of internal states \citep{prato2024large, binder2025looking, comsa2025does}, but often to a lesser degree than humans \citep{yin2023large}. \citet{ackerman2025evidence} show in a related vein that models can sometimes anticipate what answers they will give and use that information to demonstrate self-knowledge. \citet{park2026fine} also demonstrate that LLMs can be trained to introspect on their own knowledge and reference internal knowledge during generation rather than exhibiting superficial alignment. Such insights form an interesting basis for further study, which may look toward more systematically characterizing such abilities of LLMs. For example, it is of interest to understand the extent to which LLM-based systems can leverage their recognition of insufficient or outdated internal knowledge to improve at related abilities such as abstention or the selection and evaluation of different tools \citep{kumaran2026causal}. Such behavior mirrors the phenomenon of \textit{metacognitive control} in psychology.

\paragraph{Introspection: Monitoring and Reporting Internal States.} 
Separate from recognition of self-knowledge, a parallel strand of work has inspected models’ ability to introspect on internal states. Introspection is defined as the process of acquiring or accessing knowledge that originates from internal states and is not accessible to external observers \citep{binder2025looking}. Work in this direction is likewise sparse, but has taken on a few distinct directions. First, \citet{li2026language, yalon2026indications} study the ability for models to monitor and report their own belief states based on exemplars, finding that both large (up to 70B parameters) and small models ($\sim$8B parameters) often perform above chance at such tasks and can even control activation patterns in some settings. \citet{yalon2026indications} present causal evidence for the ability, and \citet{li2026language} further show that the space spanned by the directions of neural activations able to be reported or controlled in such a fashion is of significantly lower dimensionality than a model’s neural space, suggesting ability to monitor only a subset of activations covered by this ``metacognitive space.’’ Still, this capability appears to be partially mediated by the number of in-context exemplars used for the monitoring and reporting tasks, the semantic interpretability of the studied neural directions, and the amount of variance explained by such directions. A second direction of study examines whether models can identify the presence of concepts injected into activations \citep{hahami2025feeling, lindsey2026emergent}; it is shown that models can do so in certain scenarios, indicating a degree of functional introspective awareness of internal states. However, this ability is fragile. A final avenue of study specific to knowledge of language is presented by \citet{song2025language}, who, in contrast, provide negative conclusions regarding models’ introspective ability. We refer the reader to that paper for further discussion of potential explanations. More generally, the ability for models to introspect on internal states is highly context-dependent, with the factors governing it not yet fully elucidated, and presents important implications for the robustness of oversight mechanisms.

\paragraph{Behavioral Self-Awareness.} Models’ capacity to accurately describe learned behaviors without explicit examples or supervision is another important metacognition-derived capability \citep{betley2025tell}. Limited work suggests models trained to exhibit specific behaviors can articulate signatures of self-awareness across diverse tasks and contexts, and that such awareness is acquired early on in training and can be mechanistically localized to a single steering vector in a model’s activation space \citep{bozoukovemergent}. Other related investigation includes examination of introspective awareness, defined as knowledge of identity, capabilities, and objectives \citep{passaro2025awareness}, which can be improved by collaborative reasoning among multiple agents.

\paragraph{Reasoning.} The ability for reasoning models to engage in reflective and metacognition-like behaviors \citep{dong2025towards, gandhi2025cognitive, du2026latent, fu2026reefbench, kim2026reasoning, xie2026mitigating} has prompted questions regarding whether the acquisition of reasoning capabilities is accompanied by associated gains in metacognitive performance; this has led to an active line of inquiry into the connections between metacognition and reasoning in LLMs.\footnote{For example, \citet{chen2026learning} find that improving a model’s self-verification ability leads to improved reasoning performance.} To this end, several studies have provided evidence for limited metacognitive capabilities in LRMs, finding that such models struggle to leverage metacognitive information to engage in monitor or control behaviors \citep{li2026towards} and that enhanced reasoning capability can act in opposition to metacognitive sensitivity---demonstrating improved metacognition is not an automatic byproduct of better reasoning \citep{shatrade}. \citet{marjanovic2025deepseek} show that DeepSeek-R1 is incapable of simple metacognitive monitoring tasks such as detecting the length of its own reasoning traces. Despite this,  distilled variants such as DeepSeek-R1-Distill-Qwen32B and DeepSeek-R1-Distill-LLaMA-8B have been observed to maintain reliable internal estimates of relative position in reasoning, suggesting a fundamental difference in metacognitive behavior and potentially also in the acquisition of such abilities \citep{eisenstadt2025overclocking}. Lastly, the impact of metacognition on reasoning efficacy is explored to some degree by works such as \citet{lu2026auditing}; such study reveals that chain of thought faithfulness to the final answer often diverges in settings where models exhibit weak metacognitive sensitivity. It is also observed that the impact of reflective behaviors on model confidence is often minimal, suggesting a lack of metacognitive grounding. These findings underscore the need to move beyond surface-level alignment in LRMs. Moreover, metacognition remains a relatively understudied component of LRM research, with only a small portion of related papers considering metacognitive aspects \citep{kargupta2025cognitive}.

\paragraph{Task-Specific Findings.} Finally, a range of recent studies have yielded insights regarding the metacognitive skills of models in task-specific functional contexts. This includes assessment of models’ ability to: name and strategically apply skills to math reasoning tasks via guided prompts \citep{didolkar2024metacognitive}, deploy metacognitive strategies to perform knowledge editing \citep{fan2026towards}, self-reflect during essay writing \citep{Uzwyshyn2024}, introspect on generation temperature \citep{comsa2025does, song2025privileged}, perform metalinguistic analysis \citep{begus2025large}, self-assess and revise positions in adversarial debate \citep{kim2025objexmt, prasad2025two}, and execute system prompt learning \citep{pajo2025system}. Findings regarding these metacognition-derived capabilities are generally mixed across models and task contexts. Nevertheless, these settings constitute a rich testbed for probing and evaluating LLMs’ metacognitive abilities in a manner more representative of the diversity of human metacognitive faculties.

\paragraph{Impact of Model Size, Family, and Performance.} Existing works generally uncover a consistent association between model size and metacognitive ability \citep{DMC, lindsey2026emergent, park2026fine}, regardless of whether the focus of study is metacognitive efficiency, metacognitive sensitivity, or introspective awareness. This suggests metacognitive behavior may naturally emerge alongside the enhanced performance and generalization of larger models. Despite this, such findings are supported by only a limited collection of datapoints, obtained from either small, sub-9B models or large proprietary models of undisclosed size, and from specific, disjoint task settings across model scales; it is unclear whether the observed metacognitive patterns persist in moderately-sized LMs and in broader task settings and model families, including proprietary LLMs. Additionally, as noted by \citet{lindsey2026emergent}, such trends can be complex and further sensitive to other factors such as post-training strategy, which we discuss below.

Model family presents another dimension of variation affecting metacognitive analysis of LLMs. \citet{cacioli2026llms} find that metacognitive efficiency varies substantially across models even when task performance is similar, that models with similar calibration level (measured via expected calibration error) likewise exhibit distinct gradations in metacognitive efficiency, and that the impact of temperature on metacognitive sensitivity and efficiency can vary markedly by model family (e.g., Mistral and Gemma 2 exhibit reverse trends according to generation temperature). \citet{park2026fine} further demonstrate that among models of similar sizes but from different families, clear rankings according to metacognitive performance can be derived.

The use of reasoning models can introduce additional complexity that causes trends to diverge from those observed with non-reasoning LMs. In particular, LRMs with better overall performance do not necessarily exhibit the strongest metacognitive sensitivity \citep{shatrade}, and in some model families the smallest reasoning models exhibit higher metacognitive sensitivity than their larger counterparts. \citet{shatrade} attribute this to the profuse use of intermediate reasoning steps in larger models, which helps performance but increases the difficulty of accurate self-evaluation; however, their study focuses exclusively on digit manipulation tasks. On the other hand, \citet{advani2026small} study the task of metacognitive reflection and suggest that there exists a ``capacity threshold’’ of at least 7-9B for reasoning models under which metacognitive prompting interventions can be harmful. They find that while small models can identify and correct some reasoning errors via such reflection, they struggle to avoid introducing new errors during this process and often instead enter into hallucinations or logical leaps---concluding that metacognition leads to confusion without sufficient model size or capacity. It is worth noting, however, that this work defines metacognition from a conceptual standpoint that is less grounded in technical definitions seen in psychology. \citet{huang2026vulnerability} present similar findings in a non-reasoning setting that smaller models are more resistant to metacognitive prompting; yet, this may be simply due to weaker instruction-following capacity.

More generally, the lack of consistency in evaluation approach, task domain, and experimental setup (prompt, confidence elicitation approach, etc.) makes it difficult to reliably compare and validate findings and trends regarding models’ metacognitive faculties. This remains an open area for systematic, comprehensive investigation. 

\paragraph{Impact of Post-Training.} The impact of post-training procedure on metacognitive ability is not yet well-established. \citet{lindsey2026emergent} provide preliminary evidence, in addition to their commentary on the impact of model size and performance, that introspective strength is sensitive to post-training approach. \citet{cacioli2026llms} also directly compare corresponding instruct and base models from the Llama 3 family, finding a domain-specific effect in terms of the impact of post-training on metacognitive efficiency. Namely, while both variants exhibited nearly identical meta-$d'$ and $d'$, the M-ratio was slightly decreased for the instruct model, an effect more pronounced in the scientific domain versus art, literature, history, politics, and geography. They conclude that RLHF may specifically degrade metacognitive efficiency on STEM tasks. Additional analysis reveals that across different temperatures, while meta-$d'$ and $d'$ of base models remains fairly stable, instruct models exhibited a dissociation in which one of the two quantities increased or decreased while the other was stable (pattern specific to model family). Such findings point to a complex dynamic between post-training and metacognitive ability which remains to be fully profiled, despite its importance.

\paragraph{Impact of Sampling Temperature.} Generation temperature can have a measurable impact on resulting metacognitive observations. While the impact of temperature is well-recognized in the calibration literature,\footnote{That is, different temperatures can lead to different confidence estimates, especially if using sampling-based confidence estimation \citep{selfcheckgpt, chen-mueller-2024-quantifying}.} complementary findings for type 2 tasks are only just beginning, with \citet{cacioli2026llms} showing that temperature manipulation leads to divergence between metacognitive sensitivity and confidence policy for instruction-tuned models. It is observed that if a model’s metacognitive efficiency is too low, calibration via temperature adjustments cannot meaningfully address underlying metacognitive deficits, with confidence signals remaining insufficiently informative. On the other hand, base models which do not exhibit significant changes in confidence or sensitivity as a result of temperature setting do not seem to reflect this trend. The veracity and generalizability of such conclusions remain to be studied, warranting further investigation into the role of temperature in shaping demonstrated metacognitive performance of LLMs.

\paragraph{Impact of Confidence Estimation Method.} A potentially significant source of inconsistency among existing metacognitive evaluations of LLMs is the methodology used to estimate model confidence \citep{DMC}. For example, response-level confidence may be estimated via sampling consistency \citep{selfcheckgpt, becker2024cyclesthoughtmeasuringllm, chen-mueller-2024-quantifying, kaur2024addressing, xiong2024can}, semantic variability \citep{meister-etal-2022-high, kuhn2023semantic,  grewal2024improvinguncertaintyquantificationlarge, nikitin2024kernellanguageentropyfinegrained}, token probabilities \citep{kadavath2022languagemodelsmostlyknow, duan-etal-2024-shifting, Huang_2025}, or self-reported confidence scores \citep{cape, tian-etal-2023-just, hou2024decomposinguncertaintylargelanguage, yadkori2024believebelievellm, yang2024verbalizedconfidencescoresllms, zhao-etal-2024-fact}, among other approaches. While systematic study of the impact of this decision is largely absent, we summarize a few relevant insights from recent work. First, \citet{dai2026rescaling} study the impact of verbalized confidence scale manipulations on metacognitive sensitivity, finding across leading open-source and proprietary LLMs (e.g., GPT-5.2) in long and short-form tasks that confidence scale design directly impacts the quality of self-reported confidence scores and thereby also metacognitive performance. In particular, they observe that verbalized confidence values elicited in the typical 0–100 range are highly sparse and discretized, with over 75\% of responses concentrating on three values, and that alternative use of a 0–20 scale markedly and consistently improves metacognitive efficiency. To bypass this issue, \citet{DMC} leverage token log-probabilities to estimate model confidence, focusing on sub-9B open-source models in short-form QA tasks. They find that with use of such continuous confidence estimates, M-ratios approximately ranging from 0.85–1.05 are obtained, in contrast to M-ratios of approximately 0.62–0.92 obtained with the use of self-reported confidence scores \citep{dai2026rescaling}. Further evidence for the variability of metacognitive evaluation results with confidence method are provided by \citet{DMC}, whose proposed SDT analog for LLMs exhibits similar sensitivity.

\paragraph{What Metacognitive Metrics Capture that Calibration Metrics Don't.} Theoretically and empirically, analysis of metacognitive performance offers additional insights regarding the quality of model confidence signals which may not be revealed through typical calibration-based evaluation alone. Consider the setting in which multiple models appear to be well-calibrated according to metrics such as ECE \citep{guo2017calibration}, Brier Score \citep{glenn1950verification}, or AUROC \citep{steyvers2025metacognition}.\footnote{Such confidence metrics can be categorized into three tiers depending on their relevance for metacognitive assessment \citep{cacioli2026llms}: ``tier 1'' measures such as ECE and Brier Score are known to conflate sensitivity (how well correct and incorrect responses are separated) with bias (overall confidence level) and vary with discretization scheme, and ``tier 2'' measures such as AUROC and phi correlation are known to confound type 1 performance despite capturing ranking quality. Within these, the Brier score improves on ECE by decomposing into reliability, resolution, and uncertainty, but it does not separate the model’s discriminative capacity from its metacognitive sensitivity when controlling for that capacity. Likewise, the AUROC improves on ECE by measuring ranking quality, but it conflates task performance with the quality of metacognitive monitoring over that performance. On the other hand, ``tier 3'' measures include meta-$d'$ and M-ratio, which control for type 1 sensitivity and thereby isolate metacognitive sensitivity/efficiency.} In this case, metacognitive sensitivity specifically distinguishes among models which ``know what they don't know,'' which can be important for model selection in confidence-critical downstream settings and impact human-AI interactions. For example, if two models with the same ECE exhibit different $d'$ and meta-$d'$ behavior, and one model is to be deployed to a downstream system for selective prediction tasks, then the model with higher $d'$ (task performance) but lower M-ratio (metacognitive efficiency) would underperform relative to the model with lower accuracy but higher metacognitive efficiency. This tradeoff is not reflected by ECE alone. \citet{cacioli2026llms} cite such conceptual examples along with empirical evidence that AUROC and M-ratio yield fully inverted model rankings, to posit that confidence-dependent applications should therefore prioritize high M-ratio models. Metacognitive evaluation of LLMs can therefore provide a complementary lens with which to assess model quality. Yet extending such evaluation to open-answer tasks---a key benchmark setting for LLMs---remains unexplored.

\paragraph{Domain-Generality vs. Domain-Specificity of LLM Metacognition.} Whether human metacognition relies on a domain-general resource shared across different tasks or domain-specific evaluative processes has long been contested \citep{morales2018domain}. Parallel questions of interest may be formulated for LLMs. That is, are metacognitive behaviors of LLMs mediated by a single ``global’’ mechanism or governed by individual, domain-specific processes? Is the metacognitive performance of LLMs---measured via metacognitive efficiency, for example---consistent and generalized across domains? Does domain-specificity depend on the particular metacognitive skill being assessed? To the best of our knowledge, no work has yet considered the plausibility of the first or third questions. However, \citet{cacioli2026llms} supply early evidence in support of the second question that metacognitive efficiency of LLMs is, in fact, domain-specific, with different models exhibiting varied performance across task domains.

Overall, the findings highlighted in this section emphasize the importance of systematically evaluating the meta-cognitive abilities of LLMs in diverse experimental and task settings. While LLMs appear to exhibit fundamental metacognitive limitations,  complicated by factors such as post-training and confidence estimation approach, substantial opportunities remain for further investigation, particularly in understanding how these abilities can be measured, elicited, and improved, with important implications such as those discussed in \S\ref{sec:importanceofmetacognition}.

\section{Giving LLMs Metacognitive Abilities} \label{sec:endowingmetacognition}
Building on prior observations of metacognition in LLMs, we now examine methods for implementing and improving metacognitive abilities of LLMs, LRMs, and agentic systems. These approaches span architectural and modular designs, prompting-based techniques, and training strategies (e.g., use of self-generated signals during RL \citep{kim2025meta, yeom2026epicar}), among others. Notably, the prospect of encoding metacognition into AI systems predates the emergence of LLMs, with early methods leveraging metacognitive principles to improve robustness of intelligent systems \citep{schmill2008role}, synthesize general-purpose ``meta-models’’ \citep{caro2014design}, and translate human metacognitive abilities into AI analogs \citep{shakarian2026toward}, in addition to other goals \citep{wright2025treasure}. We refer the reader to such works for details on broader efforts toward artificial metacognition.

\subsection{Implementations of Metacognition in LLMs} \label{subsec:implementationsofmcinllms}
Current efforts to instantiate metacognitive functions in LLMs generally utilize specialized frameworks or architectures.

\paragraph{Frameworks.} Some studies have aimed to draw directly from metacognitive frameworks presented in cognitive psychology to develop analogous mechanisms for metacognition in LLMs. For example, \citet{yan2025position} draw inspiration from dual-process theory \citep{kahneman2011thinking} to propose a framework for metacognition-assisted reasoning in LLMs that involves self-awareness, monitoring, evaluation, and regulation. Pangu Embedded \citep{chen2025pangu} is a similarly inspired framework that adapts from dual process theory to implement system 1 (fast, intuitive) and system 2 (slow, deliberative) processing for reasoning-optimized LLMs, combining use of distillation fine-tuning, reinforcement learning, and metacognitive prompts to confer models with more nuanced, efficient, and versatile reasoning. \citet{oh2025monitor} propose the Monitor-Generate-Verify (MGV) framework, which serves as a computational translation of metacognitive theories by \citet{flavell1979metacognition} and \citet{nelson1990metamemory} and incorporates explicit monitoring to capture metacognitive experiences prior to generation as well as retrospective monitoring and verification. This approach is concretely implemented by \citet{oh2025before}, who demonstrate its benefits toward effective reasoning. More generally, \citet{scholten2024metacognitive} present a theoretical framework to explain the presence of five symptoms of metacognitive deficiency in LLMs as a byproduct of limited monitoring and control capabilities. \citet{conway2024toward} further posit that equipping LLMs with human-like metacognitive abilities requires implementation of human-like procedural memory. 

Other frameworks directly design methods to implement metacognitive functions for LLMs in specific task settings. For instance, another framework targeted toward reasoning tasks is Meta-R1 \citep{dong2025meta}, which introduces explicit metacognitive capabilities into reasoning-equipped LLMs and demonstrates consistent improvements in performance and efficiency across models and tasks. To improve mathematical reasoning, \citet{huang2025towards} present a modular framework named MetaMath-LLaMA which combines a transformer-based metacognitive subtask scheduler with symbolic parsing and symbolic-neural computation. Regarding awareness of knowledge boundaries, \citet{park2026fine} propose Evolution Strategy for Metacognitive Alignment (ESMA), which is a framework to strengthen the alignment between models’ internal knowledge and explicit behavior. A framework to simulate metacognitive learning with self-awareness, boundary monitoring, and reflective thinking to step toward human-like knowledge editing is introduced by \citet{fan2026towards}. Lastly, \citet{li2025adapting} propose a method for metacognitive test-time reasoning to enable vision LMs (VLMs) to self-reflect and learn adaptively during inference. 

\paragraph{Architectures.} To implement metacognitive functions in LMs, a few works have also proposed specialized model architectures. \citet{aviss2025state} introduce the State Stream Transformer (SST) architecture, finding that it can exhibit enhanced reasoning capabilities and emergent metacognitive behaviors, which they attribute to higher-order processes of planning and regulation. On the other hand, \citet{jha2025thinking} present SAGE-nano, a 4B parameter model which extends the transformer architecture with specialized metacognitive components that enable inverse analysis of forward reasoning. Such architectures generally maintain or enhance reasoning performance and can improve model transparency. A related innovation in optimization is the Metacognitive Introspective Reward Architecture (MIRA) proposed by \citet{zhang2025mira}, which utilizes a two-tier optimization loop to drive policy learning while incorporating metacognitive supervision to monitor learning dynamics. 

More generally, much of the effort in this space focuses on implementing metacognitive functions to assist reasoning-centric applications of LLMs; while important, empirical support for these methods can be limited, and it can be valuable to devise similar methods for broader use cases.

\subsection{Metacognition for Reasoning Models} \label{subsec:metacognitionforreasoningmodels}
A growing body of research has examined how metacognitive mechanisms can be incorporated into LRMs. \citet{ha2025aha, wang2026teaching} find that LRMs already exhibit latent reflective behaviors commonly associated with expert metacognitive reasoning, such as self-critique and reflection, motivating further work to regulate and amplify these tendencies. Inspired by this, several prompting frameworks have been developed to explicitly elicit structured metacognitive reasoning. \citet{shakoo2025dynamic} introduce the Dynamic Cognitive Orchestrator (DCO), a two-stage framework that first generates a problem-dependent reasoning strategy and then executes it, yielding strong performance on complex reasoning tasks. Similarly, \citet{fang2026probabilistic} propose the Probabilistic Chain-of-Evidence (PCE) framework, which integrates uncertainty quantification into the reasoning process via prompt engineering, consistently outperforming classical CoT on metrics such as hallucination rate. \citet{li2026towards} adopt a complementary emergent approach, dynamically adjusting thinking strategy through separate planning and solving models to improve mathematical reasoning. A parallel line of work focuses on metacognitive reuse at inference time: \citet{yang2024buffer} first introduced Buffer of Thoughts (BoT), which draws on distilled “thought-templates” from prior problem-solving processes to improve performance and generalizability. \citet{didolkar2025metacognitive} extend this idea to reasoning models, integrating a “behavior handbook”—a record of behaviors curated from previous traces—that reduces token usage while maintaining accuracy. \citet{cheng2026prism} apply a similar principle within MCTS, using a shared memory of heuristics and fallacies drawn from metacognitive reflection over prior search trajectories to guide subsequent reasoning branches, halving trajectory requirements while improving accuracy. \citet{xiang2026thinking} also target inference-time efficiency, monitoring reflection signals during generation to evaluate whether the current CoT is sufficient to terminate, reducing reasoning length by almost 35\% with minimal accuracy loss.

On the training side, researchers have developed methods to embed metacognitive capacities within reasoning model parameters, broadly falling into two categories: fine-tuning on augmented traces, and integrating self-generated signals. \citet{ha2025aha} address the problem of “overthinking”—where models generate repetitive reasoning—by decoupling reasoning and control into independently optimized components, improving accuracy and reducing token usage. \citet{wang2026teaching} apply SFT and RL on self-critiqued reasoning traces to improve both accuracy and reflection quality; \citet{sun2025cog} similarly fine-tune on self-corrected traces, adding a hierarchical decomposition step for harder problems. \citet{li2026towards} extend their prompting-based approach through SFT and RL on traces augmented with difficulty assessments and planning decomposition. \citet{didolkar2025metacognitive} similarly incorporate their “behavior handbook” into an SFT objective, achieving improved token efficiency and accuracy. The second category integrates self-generated signals into the training objective itself. \citet{kim2025meta} fine-tune LRMs by aligning predicted statistics about forthcoming generations—such as pass rate and solution length—with actual outputs, significantly improving accuracy and out-of-domain generalization. \citet{yeom2026epicar} instead incorporate self-signals regarding answer correctness, improving both accuracy and calibration—a property of particular importance for LRMs deployed in mission-critical settings. A third form of approach bypasses reasoning trace supervision entirely: \citet{leong2026finding} show that LRMs possess latent reasoning beliefs that track their own reasoning traits, and leverage this by fine-tuning on metacognitive QA pairs that reshape the model’s self-concept toward desired behaviors—matching stronger baselines at lower cost. \citet{zhao2026roi} takes a complementary angle, training models to estimate the expected value of additional reasoning before generation and enabling principled decisions about whether to invest compute in a given problem. \citet{li2026enhancing} propose mid-training on pairwise comparisons between reasoning traces to improve the reasoning-metacognition tradeoff. Finally, \citet{li2025core} take an introspective approach with CoRE-Eval, analyzing the geometry of hidden-state trajectories during reasoning to identify cyclical patterns of redundant deliberation, enabling training-free early termination that reduces reasoning length and improves accuracy.

\subsection{Metacognition for LLM Agents} \label{subsec:metacognitionforllmagents}
A third line of inquiry concerns the implementation of metacognitive capabilities into agentic systems to increase their practical utility and tackle their limitations \citep{zhang2025agents}. In particular, while LLM agents have gained traction in a variety of applications, they continue to exhibit deficiencies such as rigid task chaining, shallow reasoning, brittle memory retention, becoming trapped in unforeseen loops, or encountering unrecoverable failures, which can lead to user frustration and reduced trust \citep{xu2025agentic, alva2026agentic}. In this subsection, we discuss the works which have pursued research in this direction, categorizing them according to the specific task setting or metacognitive capability studied.

\paragraph{Learning, Introspection, Adaptation, and Self-Improvement.} Modern agentic systems encompass a wide range of structures, including use of hybrid architectures combining dynamic goal formulation and memory with natural language reasoning and environmental feedback \citep{Chase_LangChain_2022, yao2022react, Significant_Gravitas_AutoGPT}. The incorporation of metacognitive monitoring and control mechanisms presents opportunities to further improve their functionality. To this end, a number of works have developed metacognitive frameworks for self-improvement, self-prediction of competence and learning progress, improved strategy selection, and monitoring of ``thought’’ processes and actions by LLM agents \citep{toy2024metacognition, gaven2025magellan, xu2025agentic, hou2026learn, VALIENTE2026108131}. Such works generally demonstrate that augmenting agents with metacognitive capabilities enables improved task success \citep{xu2025agentic}, more effective learning in open-ended goal spaces \citep{gaven2025magellan}, more systematic reflection and use of procedural knowledge \citep{hou2026learn}, more iterative and adaptive strategy selection in diverse contexts \citep{toy2024metacognition}, better ability to handle novel scenarios \citep{VALIENTE2026108131}, strengthened subtask decomposition skills \citep{wang2024devil}, and automatic generation of agentic workflows \citep{zhou2025agentic}.

\paragraph{Multi-Agent Systems.} Similar to single-agent settings, metacognitive methods can improve the capabilities and utility of systems in multi-agent contexts. Efforts in this direction include the development of training-based or modular instantiations of ``meta-thinking’’ \citep{bilal2025meta, wan2026rema} and the implementation of fine-grained error detection and self-correction mechanisms \citep{shen2025metacognitive} for multi-agent systems. It is observed that such approaches can address known limitations of multi-agent systems, including error propagation, limited flexibility and reliability in complex or high-stakes contexts, and shortcomings in reasoning, LLM-as-a-Judge, and mathematical tasks, among others.

\paragraph{Reasoning.} Reasoning in LLM agents is another use case which has received attention. While most work to enable metacognition-assisted reasoning in LMs focuses on reasoning models directly (\S\ref{subsec:metacognitionforreasoningmodels}), some studies have devised agent-specific methodologies, including AutoCrit \citep{sang2025autocrit}, which integrates agents for critique, reasoning, and monitoring in a feedback loop to mitigate inconsistencies and error propagation; RefRea \citep{mai2026refrea}, which utilizes a metacognitive module to compare reference and reasoning model actions and guide outcomes via reflection and environment-responsive adaptation; and \citet{light2026deep} which uses scaffolded in-context meta-reasoning examples.

\paragraph{Collaboration.} Additional work has investigated the contribution of metacognitive mechanisms toward improved agent collaboration. For example, \citet{zhang2026metamind} draw upon psychological theories of metacognition and theory of mind to emulate human-like social reasoning in multi-agent settings, while \citet{yang2026adaptive} present a paradigm for human-agent collaboration that leverages judgments by a metacognitive policy to determine when to seek assistance from a human expert.

\paragraph{Memory.} As LLM agents rely on accumulated memory to tackle long-horizon and decision-making tasks, memory management is another metacognition-informed capability which presents opportunities for improvement. In this direction, \citet{liang2026learning} highlight that many agentic approaches rely on fixed memory representations at a single level of abstraction, which can hinder generalization. Thus, they propose to treat memory abstraction as a learnable skill, and introduce a metacognition-inspired approach which trains a separate model via DPO to learn to distill knowledge into representations with suitable format and at an appropriate abstraction level for reuse.

\paragraph{Search, Retrieval, and Tool Use.} A final application setting which has been explored is the use of metacognitive monitoring or reflection to better allocate resources across search, retrieval, reasoning, and other tools in agentic applications. For example, \citet{sun2026deep} introduce a monitoring framework for deep-search agents to separate execution-level control from task-level reasoning, improving the robustness and practicality of such agents. Similarly, \citet{chen2026retrieve} present a metacognition-inspired framework to enable agents to dynamically assess when to reason with existing knowledge or seek new evidence, which helps to eliminate redundant retrieval. \citet{musumeci2025robodata} also leverage metacognition-derived self-reflection mechanisms to enable agents to approach knowledge-based QA tasks in a more structured fashion and dynamically orchestrate query answering. Another related direction concerns temporal awareness, raising questions such as how an LLM might know that its knowledge is outdated to verify more recent (external) sources.

\section{Metacognitive Methods to Improve Capabilities of LLMs}  \label{sec:applyingmctoimprovecapabilities}
So far, our discussion has focused on direct observation and improvement of metacognitive capabilities of LLMs. A direction which is similarly worthwhile, especially toward realizing diverse downstream benefits of metacognition-like processing (\S\ref{sec:importanceofmetacognition}), is the devision of functionally metacognition-grounded methods to improve and extend the capabilities of LLMs. Efforts in this direction can be broadly categorized by whether they aim to improve performance in specific task settings or simply improve the efficacy and efficiency with which learning occurs.

\subsection{Improving Task Performance}
We discuss the various metacognition-inspired approaches aimed at boosting the performance of LLMs in diverse tasks.

\paragraph{General Performance.} As metacognitive processing is broadly applicable across domains, one class of work consists of methods aimed at improving the general performance and problem-solving capabilities of LLMs. Typically, models undergo post-training (e.g., instruction tuning) to improve adherence to human preferences and instructions and acquire strong zero-shot capabilities. Along these lines, some work has shown that using instructions aligned with models’ cognitive processing or inference-time metacognitive prompts inducing self-reflection and self-evaluation can drive up problem-solving and reasoning abilities across multiple benchmarks \citep{renze2024self, wang2024metacognitive, bai2025mp, zhang2025remembering}. Prompt designs capturing different types of metacognitive behavior contribute differently to resulting improvements \citep{renze2024self, metafaith}. Additionally, the ability for models to detect and explain wrong predictions in QA settings can be improved via multi-stage fine-tuning in which the model acquires experience by encountering its earlier predictions; this lends toward self-validation in a metacognitive fashion \citep{longo2025eliciting}. Yet, fine-tuning may not be required to improve the self-correction capabilities of LLMs: \citet{tan2025tuning} propose the use of sparse sub-networks to capture decision processes toward better introspective monitoring. Ample space remains to further explore how we can construct and leverage metacognitive methods to improve general capabilities of LLMs; recalling discussion of the domain generality vs. specificity of metacognition in \S\ref{subsec:currentfindings}, however, it may be more effective to focus on setting-specific instantiations. 

\paragraph{Confidence Calibration.} The ability to assess and communicate one's own uncertainty is a defining characteristic of metacognition. Yet while LLMs can maintain internal uncertainty signals, their expressed confidence is often misaligned with both accuracy and their internal uncertainty. To this end, some works have sought to devise metacognitive approaches to improve models’ ability to signal their confidence to human users in a reliable and calibrated fashion. In terms of the factuality-based alignment between confidence and accuracy, \citet{steyvers2025improving} leverage supervised fine-tuning (SFT) to strengthen models’ ability to assign calibrated confidence scores to answers and discriminate between two possible answers based on anticipated correctness. They observe that multi-task training is necessary for effective generalization of this skill. On the other hand, another strain of work has investigated metacognitive methods to strengthen the faithful alignment between models’ expressed and intrinsic confidence (i.e., ``faithful calibration’’). \citet{metafaith} show that metacognitive system prompts that directly tell a model it has strong metacognitive sensitivity can lead to consistent, moderate improvements in faithful calibration in a generalized fashion. Building upon this, \citet{rlmf} demonstrate that the use of metacognitive performance as an additional training signal during RL to scale advantage scores and preferentially refine completion rankings can lead to strong gains in faithful calibration while simultaneously conferring models with improved ability to self-predict performance in a point-wise fashion. This reinforcement learning with metacognitive feedback (RLMF) paradigm can outperform standard RL and generalizably improve performance at metacognitive monitoring and target tasks. This contrasts the SFT-based findings regarding generalizability by \citet{steyvers2025improving}, suggesting the presence of more complex dynamics governing the generalization of metacognitive capabilities---a direction of study which is unexplored as of yet. Different types of metacognitive tasks (e.g., expressing confidence to comparing question difficulty) exist, and it is possible that some forms of generalization are easier to master than others.

\paragraph{Hallucination Reduction.} The ability to self-identify errors and detect and express sources of uncertainty is characteristically well-suited for application toward hallucination mitigation. Preliminary work indeed suggests that endowing models with metacognitive capabilities---for example, by using average metacognitive error as the loss function for gradient updates during fine-tuning \citep{li2024teach}---can reduce the occurrence of hallucinations during text generation. This method is conceptually similar to the RLMF paradigm proposed by \citet{rlmf}, but uses a different measure of metacognitive performance, applies it at a different point in the training process, and focuses on fine-tuning without use of RL. It also involves a more complex design with multiple models. Nevertheless, this provides additional evidence for the value of leveraging metacognitive performance signals to inform training and optimization. Beyond training, \citet{miki2025step} also demonstrate that metacognitive prompts to enforce structured system 1 and system 2 ``thinking’’ in LLMs can reduce contextual hallucination, supporting the parallel approach of using prompts to guide models to exhibit metacognition-like behavior and subsequently improve output quality.

\paragraph{Knowledge Boundary Detection.} The ability for a model to detect and signal the scope of its own knowledge is another important skill which can inform efficient resource use, improve factuality, and boost trustworthiness \citep{prato2024large, chen2025query, lv2025whether}. Yet LLMs often experience knowledge-confidence gaps which lead to overconfident errors or uncertain truths \citep{chen2026know}. To address this, several studies have proposed the use of metacognitive methods to improve knowledge gap awareness. For example, \citet{tiwari2026self} devise metacognition-inspired reflective prompts to elicit self-evaluation and error correction in this direction, which also has applications in hallucination mitigation. A more systematic approach is introduced by \citet{chen2026know}, wherein a model’s internal signals are used to partition its knowledge space into regions of mastery, confusion, and absence and thereby inform targeted augmentation of existing knowledge. \citet{roy2025physics} further leverage physics-based constraints to improve LLMs’ self-knowledge ability and their performance at selective prediction, calibration, and response revision. 

\paragraph{Resistance to Persuasion.} In humans, a natural capability closely related to accurate confidence assessment and self-knowledge is the ability to resist persuasion when appropriate. Investigation of this property in LLMs as it relates to metacognition is limited, but preliminary evidence from \citet{huang2026vulnerability} suggests that metacognitive prompting based on the elicitation of self-reported confidence scores can, in fact, increase vulnerability to belief erosion, and that model size, family, and post-training may be important in affecting this susceptibility. Despite this, it is important to note that verbalized confidence in LLMs is known to suffer from limitations of inconsistency and unreliability. We interpret this as motivation for further research into the potential benefits of metacognition toward robustness against undue persuasion or counterfactuals, wherein alternative metacognitive prompting, training, or other methodologies could yield different conclusions, and perhaps offer an alternate path toward improving resilience to misleading or adversarially crafted inputs.

\paragraph{Interpretability.} A byproduct of improved ability to monitor and report on capabilities, knowledge, and confidence is enhanced interpretability. This view is posited by \citet{tankelevitch2024metacognitive}, who suggest that existing explainability approaches can be augmented by considering metacognition. Empirical support for this proposal is somewhat limited, but works such as \citet{plunkett2025self}, while not explicitly concerned with metacognition, highlight that models’ ability to report detailed, quantitative justifications of their decision processes can be improved through training, offering steps toward more generalized ability to describe internal operations in a way understandable to human users.

\paragraph{Reasoning for Non-Reasoning Models.} Another concrete use of metacognitive methods for LLMs is to improve the reasoning performance of models not necessarily optimized for reasoning tasks. Works in this direction generally focus on a few specific classes of approaches. Some studies present novel frameworks to adapt principles of human metacognition for LLM reasoning. For example, \citet{haque2025meta} present a metacognitive controller that enables models to dynamically select how to think: in effect, their framework uses reasoning about reasoning to identify the most suitable reasoning style (e.g., chain of thoughts, tree of thoughts, graph of thoughts) for a given input, shown to be effective for the FlanT5-Small model ($\sim$80M parameters). Other notable framework-based approaches include \citet{huang2025towards, oh2025before, ta2025mdtoc}, which focus on enhancing mathematical reasoning abilities of LLMs. These methods can improve the explainability of generated reasoning (e.g., clearer steps in solutions), beneficial toward aiding students and educators. \citet{elenjical2026think} is another noteworthy work in this space which is more psychologically grounded. A second strain of work utilizes special metacognitive prompts to enhance reasoning performance. This includes instructing models to reflect on certainty of reasoning steps and step content \citep{qian2025cognitive, fang2026probabilistic}, identify knowledge conflicts before reasoning \citep{balappanawar2025if}, and dynamically select among reasoning methods based on task requirements \citep{gao2024meta}. Such approaches often outperform chain-of-thought prompting and yield better uncertainty expression. A related technique leverages iterative reflection to induce metacognition-like processing during reasoning tasks \citep{liu2025instruct}, which can be particularly helpful for small LMs \citep{li2025reflectevo}. \citet{khandelwal2025language} show that metacognitive monitoring of performance and subsequent provision of feedback can enable progressive solution refinement, leading LMs to outperform reasoning models. Finally, metacognition-inspired analysis can be used to model LLM reasoning via influence graphs \citep{cui2025thinking}, helping to identify critical points in reasoning traces and improve reasoning efficiency and reliability. 

\paragraph{Retrieval.} As resource allocation is a key function aided by metacognition, some have devised metacognitive methods to improve the efficiency of retrieval-augmented generation (RAG) procedures \citep{li2025adaptive}. For example, \citet{chen2026retrieve} propose frameworks to dynamically evaluate exploration utility, identify coverage or relevance deficiencies, and assess when to seek new evidence or reason. This helps to eliminate redundant retrieval steps and improve context evolution and conciseness, leading to gains over text- and knowledge graph-based RAG baselines. Detection of knowledge boundaries is another metacognition-derived ability which \citet{lv2025whether} utilize via in-context learning to increase the accuracy of self-knowledge assessment prior to retrieval. General application of metacognitive functions such as monitoring, evaluation, and planning has also been used to assist retrieval-augmented LLMs to significantly outperform prior work \citep{zhou2024metacognitive}.

\paragraph{Others.} Metacognitive methods have also been used to improve other diverse abilities of LLMs. This includes metacognitive frameworks, training pipelines, prompting methods, and decision guidance. For instance, \citet{lin2025think, qin2025r, shan2025mentor} devise frameworks incorporating metacognitive self-assessment, error detection, and learning strategies to boost LLMs’ ability to identify and mitigate domain-specific risks, engage in creative planning, maintain consistency in role-playing tasks, and learn from distillation training. Toward risk mitigation, \citet{shan2025mentor} posit that metacognitive monitoring strategies can help to navigate the generalization-specialization tradeoff by bypassing the need for task-specific training while avoiding the loss of depth associated with overly general solutions. \citet{qin2025r} show that metacognitive evaluations during training can be used to provide interpretable intermediate generation states. \citet{lin2025think} also find that metacognitive learning can foster creativity and improve efficiency, enabling models to perform robotic tasks with minimal demonstrations. Beyond this, the incorporation of metacognitive knowledge during training by attaching annotations of, for example, skills required to solve a question, to training samples has been shown to help reduce catastrophic forgetting \citep{zhang2026metagdpo}. Metacognitive prompting methods have been applied toward improving  prompt engineering \citep{maxwellmeta, qiu2025mela} and models’ ability to perform causal inference \citep{ohtani2024does}, self-identify and reduce biases \citep{hills2026could}, detect sarcasm \citep{lee2025pragmatic}, and generate ontologies \citep{lippolis2024ontogenia}. Additionally, metacognitive signals (e.g., self-evaluation of competence) have been leveraged to support adaptive decision-making in tool use and pedagogical settings, enabling models to dynamically regulate when to invoke external tools and guide human learning processes in a more interpretable manner \citep{li2025adaptive,liu2026metaclass}.

Taken together, the works discussed in this section demonstrate that metacognitive monitoring and control behaviors can lead to tangible benefits for diverse downstream abilities of LLMs. Despite this, conclusions appear scattered across models and tasks, some of which are outdated.

\subsection{Improving Learning Efficacy}
We introduce metacognitive methods by which to improve the efficiency and efficacy of learning and skill acquisition in LLMs.

\paragraph{Task Monitoring.} The ability to reliably evaluate and track task progress is a fundamental aspect of efficient and effective learning. Such monitoring enables individuals to regularly compare their current state of understanding against an intended goal and adjust accordingly. However, it is often hindered in LLMs due to their poor metacognitive calibration, systematic overconfidence, and limited self-monitoring and regulation (\S\ref{subsec:currentfindings}). To this end, recent studies have begun to investigate whether explicit metacognitive supervision can serve as a solution. In the context of deep search tasks, \citet{sun2026deep} observe that agent failures often are not purely attributable to weak reasoning, but rather to a lack of mechanisms to detect when retrieval states and reasoning trajectories become unreliable. Thus, they introduce a hierarchical monitoring framework consisting of a fast consistency monitor and slow experience-driven monitor. The fast consistency monitor is responsible for ensuring the model's reasoning aligns with external evidence. If there is discrepancy, then the slow experience-driven monitor accesses memory records to generate corrective actions. This simulates a form of metacognitive monitoring that enables improved regulation of task completion progress. Similarly, \citet{roy2025physics} present a fine-tuning approach to improve models’ metacognitive calibration when learning physics-based tasks, leading to improved prediction performance and monitoring capabilities and providing further support for the value of supervisory frameworks for metacognitive monitoring. Relatedly, \citet{fan2026towards} present a framework emulating the learning process of humans with meta-memory mechanisms, enabling multimodal LLMs to better (metacognitively) monitor the scope and reliability of new information. This improves learning success against prior cognition-inspired approaches and further improves knowledge editing on multiple benchmarks.

\paragraph{Self-Critique.} Reflection is a metacognitive trait that promotes error identification and revision toward improved learning outcomes. To this end, recent work has implemented metacognitive self-critique strategies to enhance the reflective capabilities and subsequent performance of LLMs. \citet{hills2026could} demonstrate that LLMs can be primed to evaluate and convey biased decisions in their reasoning via metacognitive prompting. \citet{wang2026teaching} additionally move beyond superficial reflection behavior via a training framework that leverages high-quality reflections to construct reward signals for reinforcement learning that guide models to internalize self-correction procedures, leading to effective self-distillation. Notably, they provide evidence refuting earlier propositions that models’ inherent reflection  behaviors are truly metacognitive.

\paragraph{Strategy Selection.} The ability to make accurate judgments of what method to use and when one is better than others is another crucial feature of learning and planning that is informed by metacognition \citep{cary2002metacognition}. Existing metacognition-inspired approaches toward this mainly focus on LLM agents and are discussed previously in \S\ref{subsec:metacognitionforllmagents}. One application of this ability is toward ensemble selection, where multiple expert models are available and chosen based on a selection policy that maximizes the cumulative reward. In this direction, \citet{phammetacognitive} leverage metacognitive sensitivity to inform model selection, finding that this mechanism improves joint accuracy over individual performance and is superior to standard confidence-based selection---which often obscures metacognitive qualities that are important for a complete understanding of the reliability of models’ confidence signals and subsequent decisions regarding strategy selection (\S\ref{subsec:currentfindings}).

\paragraph{Self-Improvement.} More generally, the ability of a model to generalizably and iteratively improve its performance and capabilities on its own is a key aspect of metacognition that underpins efficient and effective learning \citep{liu2025position}. A few works investigate this by proposing frameworks for models to self-improve upon their reasoning, correction, and evolution capabilities. For instance, \citet{li2023mot} propose a  framework in which models can self-improve without additional training by ``pre-thinking’’ using high-confidence reasoning paths stored in memory, reinforcing the idea that self-improvement depends on effectively abstracting and recalling successful and relevant problem-solving. Complementing this perspective, \citet{qu2024recursive} show that self-improvement can be implemented via self-evolving mechanisms, introducing a fine-tuning framework to enable continual introspection and self-refinement in LLMs. \citet{bao2025galaxy} further implement a framework for continual and autonomous improvement that integrates metacognitive monitoring, adaptive memory, and dynamic learning; the aim is to enable agents to identify knowledge gaps, seek relevant information, and refine their internal representations over time.

\section{Applications of LLM Metacognition} \label{sec:applications}
Conferring LLMs with more human-like metacognitive capacities and behaviors can be useful toward enhancing downstream human-AI interactions. The study of such benefits is relatively limited; we summarize three primary directions.

\paragraph{Human-AI Decision-Making.} Collaborative human–AI decision-making has proven to be a powerful approach for improving task outcomes, with the metacognitive performance of LLM-based and other AI systems playing a critical role in maximizing decision accuracy. It is well-established in psychology that metacognitive ratings are important for optimal joint decisions by humans \citep{lee2025metacognitive}. In a similar vein, recent works have highlighted the complementary contribution of \textit{metacognitive sensitivity} (recall: the ability to assign confidence scores that discriminate correct from incorrect responses) in conjunction with predictive accuracy of LLM-based and other AI systems toward optimizing hybrid decision-making outcomes \citep{li2025beyond, li2025importance}. Such studies have formulated theoretical frameworks to model and evaluate the relative impact of these two properties in hybrid decision-making contexts, showing that better AI metacognitive sensitivity increases human decision performance and that agents with lower accuracy but higher metacognitive sensitivity can actually improve overall accuracy. These results underscore the importance of prioritizing metacognitive sensitivity of models to achieve superior decision outcomes. More broadly, the reliability of confidence estimates issued by AI systems is critical to helping users to calibrate their trust in such systems and optimize their consideration and use of AI advice \citep{lee2025metacognitive}. \citet{fernandes2025ai} find that use of AI tools can help mitigate systematic overconfidence in humans but sometimes reduces the precision with which humans assess their own performance. \citet{colombatto2025metacognition} further identify differences in self-confidence and response detail when users accept or reject LLM assistance in event planning settings. 

Yet as reliance on AI expands, human metacognitive skills become increasingly necessary \citep{huff2024metacognitive}. Stronger metacognitive capacities for LLMs and other AI systems does not remove the burden of discernment from human users. In fact, use of generative AI can impose multiple metacognitive demands on human users, even as explainability helps to offload some of the work \citep{tankelevitch2024metacognitive}. Several studies have observed increased ``metacognitive laziness’’ in human users who depend on LLM assistance, wherein they become less actively engaged in critical thinking and learning \citep{fan2025beware, singh2025enhancing}. To address this, some have proposed targeted efforts to increase metacognition-based AI literacy \citep{sass2025meta, bink2026can}, including tasking users with deliberately engaging in critical thinking and rational reasoning when considering LLM-generated information \citep{ladd2026skeptical}, and undergoing metacognition training with special prompts and quizzes \citep{suyou}. The timing of feedback provided to humans during AI-assisted learning can further play a role in such metacognitive judgments \citep{nadila2024impact}.

\paragraph{User Simulation.} LLM metacognition can be used to simulate human users to test AI systems for downstream settings such as teaching or therapy, in instances where it is difficult or not technically possible to effectively use real humans. \citet{borchers2026large} investigate the accuracy with which LLMs anticipate novice learners’ success at problem-solving and reflect their metacognitive models, finding that such LLM simulators tend to generate overly coherent and confident reasoning compared to real student think-alouds. To combat this weakness, \citet{zheng2025cognitive} propose Cognitive Echo, which fine-tunes LLMs on authentic learning interaction data to allow them to generate reasoning that resembles learners’ cognitive processes and strategies more closely. This enables researchers to anticipate and capture student cognitive strategies and behaviors that would otherwise go undetected. Toward identifying high-quality student agents, \citet{li2025type} introduce a pipeline to generate and filter LLM simulators by first automatically generating diverse student profiles across various attributes (e.g. demographics, personality, learning-based characteristics), then running automated scoring and evaluating a subset of them through human expert evaluation, and lastly applying graph-based score refinement across similar profiles. Besides simulating students, LLMs can also be used to simulate clients in therapy sessions. Current simulated clients unrealistically reveal and understand their thoughts and feelings, making it difficult to assess whether a model can genuinely explore a client’s internal beliefs. Therefore, \citet{kim2025can} present MindVoyager, which provides controllable levels of openness and metacognition that dynamically adapt to the progress of the therapy session, and show that this setup enables more realistic testing of LLM therapists.

\paragraph{Pedagogy.} 
As metacognition is widely recognized as a critical driver of effective learning \citep{flavell1979metacognition}, LLMs---and AI-powered tools more broadly---have the potential to serve as metacognitive tools in educational settings, by scaffolding reflective learning practices through intelligent design \citep{lee2025learning}. \citet{lee2026rewarding} introduce PedagogicalRL-Thinking, a framework that rewards the model’s thinking process and uses pedagogically-grounded prompting, yielding significant improvements in student solve rates. Other works explore interactive designs: \citet{holub2026reflecting} use multi-agent Socratic dialogue to iteratively refine metacognitive reflection questions, \citet{holmes2026llm} find that LLM-generated prompts designed around metacognitive objectives consistently outperform alternatives, and \citet{tomisu2025cognitive} reconceptualize AI as a ``Cognitive Mirror”---a teachable novice whose deliberate ignorance forces learners to externalize and monitor their own understanding. \citet{gatzia2026developing} similarly investigate how LLM-based activities can foster metacognitive development through an “Essay Analysis Assignment” in which students critically evaluate AI-generated text.

However, this promise does not always come to fruition. \citet{boettcher2026autonomous} finds that students rarely use LLMs for metacognitive tasks---only 23\% do so in engineering lab settings. Even when students do engage, the quality of interaction varies considerably: \citet{contel2025investigating} observe that many students rely on ChatGPT as a passive replacement for their own reasoning rather than using it diagnostically, while the model itself frequently fails to scaffold planning and self-monitoring. \citet{uittenhove2025metacognitive} report similar findings, showing that a well-prompted metacognitive chatbot sees low engagement across three educational contexts, with no link between engagement and learning outcomes. Notably, however, \citet{contel2025investigating} find that students who do express metacognitive awareness and use the LLM as a diagnostic tool see increased benefit---suggesting that realizing the metacognitive potential of LLMs in education requires not just capable models but careful pedagogical design around them.

\section{Reflections \& Discussion} \label{sec:discussion}

\paragraph{Why metacognition?} Metacognition is a key aspect of intelligence concerning the processes by which we monitor and control our own cognition. Endowing language models with metacognitive capacities can enable them to become more reliable, efficient, and effective at learning, problem-solving, decision-making, and planning. This can improve the performance of LLMs on downstream tasks in a more generalized fashion, increase adaptability to user needs, and extend fundamental LLM abilities \citep{gao2024meta, shan2025mentor}. It can also make LLMs more interpretable and explainable and enhance the reliability and quality of human-AI interactions.

\paragraph{Are LLMs really able to exhibit metacognition?} So far, it is clear that LLMs are not yet capable of robust metacognition \citep{tison2025human}. Indications that LLMs exhibit weak or limited metacognition (\S\ref{sec:dollmshavemetacognition}) include (1) their systematic tendency toward overconfidence and misaligned reasoning and confidence expressions \citep{leng2025taming, cash2026quantifying, shatrade}, (2) their struggle with tasks involving metacognitive judgments of knowledge, performance, and resource allocation \citep{yin2023large, binder2025looking, huff2025judgments, hwang2025can, kale2025line, zhao2026roi}, (3) their fragile introspection and revision abilities \citep{yona2024can, hahami2025feeling, prasad2025two, li2026language}, and (4) their susceptibility to redundant information \citep{scholten2024metacognitive}. LLMs also appear to differ from humans in their metacognitive capacities and behaviors \citep{steyvers2025metacognition}. Still, some observations suggest that LLMs may indeed exhibit metacognitive tendencies and that these are similar to humans, including the ability to improve at metacognitive skills via training or prompting interventions (\S\ref{sec:endowingmetacognition}, \S\ref{sec:applyingmctoimprovecapabilities}), the exhibition of metacognition-derived abilities during reasoning and reflection (\S\ref{subsec:metacognitionforreasoningmodels}), differing levels of metacognitive performance across task settings, and the seemingly disjoint relation between performance and metacognition (\S\ref{subsec:currentfindings}). Further in-depth analysis of factors such as model architecture, optimization objectives, training data and procedure is needed, along with the development of more systematic benchmarks to measure metacognitive abilities of LLMs \citep{zhang2025remembering}.

Additionally, there is ongoing debate about whether models can actually engage in metacognitive processes rather than superficially simulating them \citep{qian2025cognitive}. This mirrors broader discussion of whether models exhibit genuine understanding or merely (pretend to) imitate it \citep{ohtani2024does}, including questions about the authenticity of LLM reasoning. Consciousness is not necessarily a prerequisite for metacognition \citep{newton1995metacognition,rosenthal2000consciousness, kunimoto2001confidence, timmermans2012higher}, and it is possible for metacognition to operate without an individual being aware of it, without belief in one’s own consciousness, and without introspection \citep{kentridge2000metacognition, winkielman2012consciousness, cash2026quantifying, kaiser2026no}. Metacognition can also \textit{functionally} operate in entities such as LLMs. However, some propose that making LLMs more human-like in their metacognition may require a normative shift in tension with the goal of using them as tools \citep{tison2025human}. It is also important to be precise in use of the term metacognition, and to be careful to not conflate it with, e.g., simple reflection processes \citep{erbe2025consciousness}.

\paragraph{What are challenges to designing effective methods for LLM metacognition?} Metacognitive interventions for LLMs require careful design to be efficacious. There is conflicting evidence, for example, on the utility of meta-reflective prompts in different settings \citep{renze2024self, metafaith, fc_in_lrms, tiwari2026self}. Model-specific design may be important, as adapting strategies for one model to others may neglect model-specific metacognitive representations \citep{oh2025before}. This parallels the privileged access humans have to their own metacognition \citep{nelson1990metamemory}, with the helpfulness of different metacognitive strategies and training varying across individuals \citep{mccombs1988motivational, WINNE1996327, taub2014can, carpenter2019domain}. The choice of metacognitive approach and degree to which it is psychologically informed are other factors warranting consideration. Training-based methods remain limited, with much of the existing work relying on prompting, despite spanning diverse use cases. This raises questions regarding the extent to which metacognitive strategies should be designed for generalizability versus tailored to specific tasks, the level of complexity required (e.g., whether sophisticated frameworks are necessary), and the degree to which psychological principles should be incorporated. Alternate views on (AI) metacognition \citep{kawato2021internal} may be useful as well; for example, \citet{fitousi2025bits} reformulate metacognitive sensitivity as an information processing problem involving multiple agents, providing a different theoretical operationalization for studying confidence and metacognitive efficiency. Unreliable self-reporting and sensitivity of metacognitive estimates to confidence elicitation methodology present additional questions for analysis and design in this space. When self-reported signals are used, the distinction between prospective and retrospective judgments can be relevant \citep{siedlecka2016but}. Lastly, safety implications of metacognition for LLMs are noteworthy to consider as the field progresses. It may be valuable for alignment and metacognition advances to go hand in hand, along with attention to differences between human and AI metacognition, which can enhance oversight and collaboration, respectively. It is also important to be mindful of the apparent domain specificity of LLM metacognition, as a model being metacognitively efficient in one domain and poor in another could pose deployment risks.

\paragraph{What are risks of LLM metacognition?} Recent studies have suggested that LLMs can monitor, report on, and modulate their internal states and neural activations, and exhibit behavioral self-awareness, the ability to accurately describe or predict learned behaviors without explicit supervision \citep{li2026language}. Investigations such as \citet{betley2025tell} have also found that models can be aware of backdoor attacks and other vulnerabilities. If a model is honest, such capabilities enable it to identify problematic tendencies that can arise in its behavior (e.g., training biases, data poisoning \citep{chen2017targeted, evans2021truthfulaidevelopinggoverning, wan2023poisoning, carlini2024poisoning}). However, this also raises safety concerns, as a dishonest model could leverage its awareness to strategically misrepresent or purposefully obfuscate its true abilities and internal processes during evaluation \citep{greenblatt2024alignment, bozoukovemergent}. To this end, establishing a better understanding of the extent of LLMs’ metacognitive abilities can be critical. The autonomy associated with certain metacognitive capacities (e.g., self-directed planning or learning) must also be considered, and potentially subject to monitoring in safety- or privacy-critical contexts \citep{alva2026agentic}. It is also of interest to speak of the interplay between human and LLM metacognition. A human user who usually has good metacognition may confidently believe they can detect when AI is misleading them and therefore scrutinize outputs less carefully relative to user with less confidence in their own evaluative skills \citep{maynard2026ai}. Vigilance against the wrong threats is another risk if users are inadequately informed of a system’s metacognitive abilities and associated safeguards. Some further have posited that models capable of metacognitive processing should bear greater responsibility for actions and decisions \citep{ribeiro2023metacognition, fleig2025meta}, representing another direction for further discourse.

Despite potential risks, metacognition for LLMs can also help to improve safety and transparency in high-stakes settings by directly combating overconfidence to reduce the likelihood of systems failing to recognize when they are pursuing flawed solution paths or detrimental strategies \citep{rivera2024escalation}. Metacognitive deficits are often credited as the source of discrepancies between models’ internal uncertainty and expressed confidence, or between their internal decision-making and verbalized reasoning processes, so improved metacognition can further improve the faithfulness of models’ communication with human users, boost interpretability, and enhance our ability to detect flawed decision-making \citep{lanham2023measuring, chua2501deepseek}. Overall, the risks and benefits of metacognition require careful attention and additional consideration.

\section{Future Directions} \label{sec:futuredirections}
Metacognition in LLMs remains a nascent and exploratory field, and many directions for future research are possible given the broad scope of metacognitive abilities. We highlight a representative set of directions below, beginning with more general and foundational avenues before moving on to emerging topics and concluding with more long-term, speculative questions.

\paragraph{Measuring, Understanding, Improving, and Applying LLM Metacognition.} While LLM metacognition has wide-ranging applications, more systematic, robust, and comprehensive methodologies to measure and quantify LLMs’ existing metacognitive faculties are required to advance progress. Current paradigms to assess particular metacognitive abilities of LLMs or quantify their metacognitive skills in specific task settings lack consistency in measurement techniques, confidence elicitation strategies, task formulations, models, and other experimental factors such as temperature and dataset size (\S\ref{sec:dollmshavemetacognition}), making it difficult to assess the reliability and generalizability of resulting observations. Metacognition-focused benchmarks suffer from similar limitations in systematicity and scope and are few in number (\S\ref{subsec:measuringmetacognition}). Tasks considered can be artificial and may not reflect real-world settings. Further, metacognitive capacities of LLMs are largely neglected in general evaluation benchmarks, despite the various ways in which they impact the downstream performance and utility of LLMs and the complementary insights metacognitive metrics can provide beyond standard measures of confidence calibration (\S\ref{subsec:currentfindings}).

It is also not yet clear how we can explain the emergence of metacognition-like processing or behaviors in LLMs, nor whether such abilities manifest in a generalized or domain-specific fashion. Understanding the role of post-training procedure, training data properties (e.g., diversity, prompt structure), model architecture, and other factors on LLM metacognition represents one line of inquiry in this direction. The application of mechanistic interpretability techniques or other targeted analysis to uncover underlying mechanisms responsible for observed metacognitive behaviors presents another unexplored and fruitful avenue.

Toward improving and applying LLM metacognition, future work could investigate the devision and use of metacognition-inspired mechanisms for a wider range of models, task scenarios, and practical applications, potentially in combination with existing approaches. For example, metacognitive reward design has been considered by several works and yielded success in improving the faithfulness and calibration of LLMs (\S\ref{subsec:implementationsofmcinllms}, \S\ref{sec:applyingmctoimprovecapabilities}). The extent to which metacognition can support learning efficiency and efficacy and other fundamentally relevant functional processes also remains to be explored. More sensitive and well-calibrated metacognition in LLMs may further unlock new capacities such as self-directed behaviors and learning, curiosity-driven exploration, and creativity \citep{steyvers2025metacognition}.

\paragraph{Metacognition and Creativity.} Creativity is important for a variety of cognitive abilities and tasks, such as problem-solving \citep{bai2025mp}, robotic planning \citep{lin2025think}, and research ideation \citep{liu2026owns}. In humans, metacognitive feelings are integral to the generation, evaluation, and selection of creative ideas \citep{puente2023metacognitive}. It is also of importance in co-creative settings, which are an increasingly prevalent form of human–AI interaction wherein metacognition-supported reflection can mediate how users adapt creative strategies over time and interpret system behaviors and design intentions \citep{wang2026reflexa}. Preliminary work shows metacognitive prompts can improve the explainability and efficacy of human-AI scientific ideation systems and help researchers to examine how ideas evolve \citep{liu2026owns}. Further benefits to task performance and downstream applications may be achieved by studying creative mechanisms for LLMs based on metacognitive principles.

\paragraph{Metacognition and Self Improvement.} As metacognition is a fundamental part of learning that enables active evaluation, reflection, and adaptation, it may be key to the development of self-improving agents. Self-improvement refers to the ability to continuously acquire new capabilities and experience with minimal (human) supervision. Current approaches to instantiate this ability in LLM-based agentic systems are often rigid or lack generalization and scalability. It is suggested by some \citep{liu2025position} that this is due to reliance on purely \textit{external} metacognitive mechanisms, which depend on fixed, human-designed implementations of metacognition, instead of intrinsic metacognitive learning. Regardless, endowing LLMs with improved metacognitive capacities represents a promising direction toward more generalized and aligned self-improvement. How best to balance metacognitive responsibilities between humans and agents in this process is an open question.

\paragraph{Metacognition and Theory of Mind.} The study of metacognition in LLMs can be helpful toward understanding their theory of mind capabilities---another aspect of human cognition of recent interest for LLMs, which has implications for collaborative or social environments involving multiple agents. Theory of mind (ToM) describes the ability to attribute unobservable mental states (e.g., intentions, desires, beliefs) to others, and it is closely related to mentalizing, which refers to the application of metacognition-like monitoring to other individuals \citep{frith2012role, kosinski2023theory}.\footnote{For example, taking into account an individual’s mental states (monitoring) and using this to predict behavior (control).} A number of works have suggested that LLMs can demonstrate emergent ToM abilities \citep{amirizaniani2024can, chen-etal-2025-theory, 10.1098/rstb.2023.0499, villa-cueva-etal-2025-moments}, although findings are mixed. ToM can improve models’ ability to model user psychology, learn about users in a humanistic fashion, and generate responses that are more effective, adaptive, and contextually appropriate (e.g., in terms of style, scope, tone, persuasiveness), with further utility in pedagogical and other agentic settings \citep{NEURIPS2024_8340b085}. Notably, \citet{leer2023violation} demonstrate that metacognitive prompting can improve ToM in LLMs and reduce ToM prediction error in AI tutoring applications. Considering the functional similarity of metacognition and ToM, conferring LLMs with better metacognition may have implications for their ability to model and reason about the cognitive processes of other agents or human users.

\paragraph{The Prospect of Meta-Metacognition.} Recent evidence suggests that humans possess the meta-metacognitive ability to assess the quality of their own metacognitive judgments of confidence \citep{Mazancieux_mmc, 10.1093/nc/niad023, SCHWARTZ2026101254}. Whether LLMs possess this ability has yet to be studied and presents another interesting direction for future work toward characterizing metacognition in LLMs. Comparison between human and LLM metacognition may also provide valuable insights toward this and other broader questions, and serve as a framework for more principled accounts of LLM metacognition. 

\section{Conclusion} \label{sec:conclusion}
This paper provides the first comprehensive and up-to-date review of the current state of research and knowledge on metacognition in LLMs. We organize and unify existing work on this topic, discussing techniques for measuring and eliciting metacognition in LLMs, methods to improve and apply LLMs’ metacognitive abilities, findings and implications of work in the area, and challenges and open directions for future research. While LLMs have made remarkable progress on many NLP tasks and in diverse downstream settings, the extent to which they can display, acquire, and apply metacognitive faculties remains unclear. Further investigation is needed to understand these, as well as whether models are capable of genuine metacognition or simply simulating memorized patterns, how metacognition can facilitate other desirable behaviors and qualities, and the implications of LLM metacognition for safe oversight, deployment, and human-AI interactions. We hope this paper serves as a foundation for further exploration and discourse regarding this important topic.

\section*{Limitations} \label{sec:limitations}
In this paper, our aim is to provide a systematic and comprehensive overview of existing research on metacognition in LLMs. While we reference numerous studies related to this topic and make the best effort to cover the most relevant works to date, it is possible that some work was overlooked. Additionally, given accelerating interest in the field, there may have been new contributions produced concurrently with writing and review of this paper. We encourage readers to also consult future papers citing this one for a more updated understanding of methods and applications of metacognition in LLMs. Finally, we note that within our discussion, where model confidence is mentioned, we primarily address it through the lens of metacognition rather than providing a comprehensive treatment of calibration, which is a well-established area with numerous existing surveys \citep{huang2024surveyuncertaintyestimationllms, xia2025survey}.

\section*{Ethics Statement} \label{sec:ethics}
As this paper did not conduct experiments, engage with sensitive datasets, or employ annotators, the work itself does not have ethical implications. Nevertheless, metacognition in LLMs is relevant to confidence calibration, self-awareness, and self-improvement, which present implications for oversight, safety, and alignment that merit careful consideration as the field advances.

\bibliographystyle{plainnat}
\bibliography{references}

\end{document}